\RequirePackage{fix-cm}
\documentclass[smallextended]{svjour3}       
\smartqed  
\usepackage{graphicx}
 \journalname{Soft Computing}
\begin{document}

\title{On Randomization of  Neural Networks\\as a Form of Post-learning Strategy
\thanks{This work has been supported by the project EC AComIn (FP7-REGPOT-2012–2013-1), and by the Bulgarian Science Fund under grant DFNI I02/20.}
}


\author{K.G.~Kapanova         \and
        I.~Dimov            \and
        J.M.~Sellier 
}


\institute{K.G.~Kapanova \at
              IICT, Bulgarian Academy of Sciences, Acad. G.~Bonchev str. 25A, 1113 Sofia, Bulgaria \\
              Tel.: +359-2-9796608\\
              \email{kapanova@parallel.bas.bg, kkapanova@gmail.com}           
           \and
           I.~Dimov \at
              IICT, Bulgarian Academy of Sciences, Acad. G.~Bonchev str. 25A, 1113 Sofia, Bulgaria \\
           \and
           J.M~Sellier \at
              IICT, Bulgarian Academy of Sciences, Acad. G.~Bonchev str. 25A, 1113 Sofia, Bulgaria \\
}

\date{Received: date / Accepted: date}

\maketitle

\begin{abstract}
Today artificial neural networks are applied in various fields - engineering, data analysis,
robotics. While they represent a successful tool for a variety of relevant     applications,
mathematically speaking they are still far from being conclusive. In particular, they suffer
from being unable to find the best configuration possible during the training process (local
minimum problem). In this paper, we focus on this issue and suggest a simple, but effective,
post-learning strategy to allow the search for improved set of weights at a relatively small
extra computational cost. Therefore, we introduce a novel technique based on analogy    with
quantum effects occurring in nature as a way to improve (and sometimes overcome)        this
problem. Several numerical experiments are presented to validate the approach.

\end{abstract}

\section{Introduction}
\label{intro}
Neural Networks represent a multidisciplinary field, including e.g. neuroscience, mathematics,
statistics, computer science, engineering, and physics.  We can generally look          at the
development of the artificial neural network (ANN) field as few periods of extensive research,
starting with the first idea of a neuron by W.~McCulloch and W.~Pitts (1943) \cite{MCPitts}. In the    $80s$,
J.~Hopfield introduced his recurrent neural network,  while P.~Werbos developed            the
back-propagation algorithm - one of the most widely used to this day \cite{Haykin}, \cite{Werbos}.
The reader should note that all developed networks are based on deterministic approach.
\bigskip

ANNs try to mimic only the four fundamental elements of the biological neurons -        input,
processing, learning and output. In order to be able to generate an output,   they exploit the
interconnection principle between biological neurons. We can broadly classify them by the type
of learning: supervised or unsupervised. Through the learning (or training process),       the
weights and biases of the network are adapted. There are many strategies for learning, usually
determined by the way the values are evolved. The main important characteristics        of the
learning process are represented by two concurring passages, i.e. the capacity to use  minimum
computational resources, and to provide robustness of the system. One further step remains the
choice of the error function, or target function and the goal is to minimize the      error by
varying the weights of the ANN.

Nowadays in the standard approach, the changes in the weights are usually accomplished  during
the training process, with particularly nothing being implemented in the utilization    of the
network. Among the possible options used in the learning process, the gradient descent  method
is one of  the most common. The back-propagation algorithm proposed in \cite{Werbos}  uses the
error to propagate it through the network layer by layer until an outcome is produced.     The
backpropagation looks for the minimum value of the error function in weight space      through
delta rule or gradient descent. On the other hand we have Evolutionary Algorithms,   which are
generally directed random searches. They start from a set of random population,         slowly
converging to a solution \cite{Branke}. Another interesting alternative is represented      by
simulated annealing \cite{Kirkpatrick}, which in some situation can perform        faster than
backpropagation or genetic algorithms. By analogy with the physics problem discussed        in
\cite{Metropolis}, the strategy is based on the transition process of a solid substance   from
increased temperature to thermal equilibrium. In this context, the cooling of a      substance
becomes equivalent to minimize the cost function of an optimization problem.                In
\cite{Kirkpatrick} the simulated annealing method is achieved by substituting the cost     for
energy, and executing the algorithm by slowly decreasing the temperature values. 
\bigskip

To the best of our knowledge, the above concepts are strongly based on analogies          with
classical (or deterministic) physics. An interesting possibility is to exploit, in some sense,
quantum mechanical effects in a ANN. This was suggested for the first time in \cite{consciousness}
where biological aspects of quantum phenomena in brain activity related to the      activation
point by a nerve impulse are described. 

In this article, the action potential plays a fundamental role in information       processing
inside the brain. The authors suggested that the firing of a neuron is obtained by the  motion
of a quantum particle, in the proximity of a potential (or energetic) barrier where  typically
effects such as tunnelling occurs. Alternatives to this explanation exist such as      the one
proposed in \cite{Penrose} based on the concept of micro-tubules which are able to    maintain
a macroscopic coherent superposition. In particular, the suggested explanation described    in
\cite{consciousness} inspired us to develop a technique which mimics quantum effects in  order
to improve the set of weights of a ANN in the {\sl{post-learning}} stage. In more details, our
method aims to reinforce the reliability of a network even in the case of training failure, at
a relatively low computational cost.

The paper is organized as follows \footnotemark. In the next section we introduce          the
methodology behind the development of our proposed network. Then, in order to validate    this
novel approach, we perform a set of numerical experiments involved in the problem           of
approximating a known function. In spite of the simplicity of the proposed technique,       we
believe that it provides a further chance for the network to escape from local minima,   local
optima or saddle points \cite{Pesky} at a reasonable computational burden.
\footnotetext[1]{The reader should note that in this paper we interchangeably use    the words
{\sf quantum}, {\sf random} and {\sf noise} having in mind the same meaning. The same  applies
for the terms {\sf classical} and {\sf deterministic}.}

\section{Formulation and Methodology}
\label{Formulation and Methodology}
Nowadays, Neural Networks come in a great variety of ways - classification, data     analysis,
dimensionality reduction, etc, and thus there are many different implementations of     neural
networks. Among them we have, for example, the perceptron \cite{Rosenblatt}, the    multilayer
feedforward network  \cite{Bishop}, the probabilistic network \cite{Specht}.       Interesting
alternative based on fuzzy logic can be found in \cite{GENEFIS},\cite{PANFIS}.   In this work,
we focus on multilayer feedforward network, which consist of many neurons, each of them  fully
connected to every neuron in adjacent forward layers, although the technique is not limited to
this particular implementation.

\begin{figure}
\centering
\begin{minipage}{0.75\textwidth}
\begin{tabular}{c}
\includegraphics[width=0.75\textwidth]{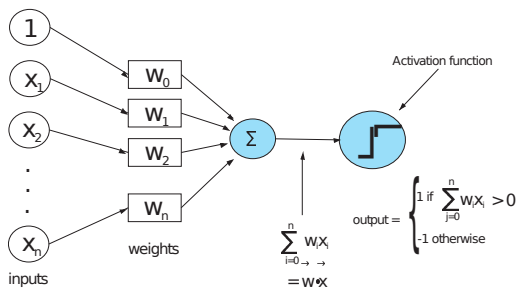}
\end{tabular}
\end{minipage}
\caption{Sketch of a typical neural network architecture.}
\label{fig:neuron}
\end{figure}

Figure \ref{fig:neuron} describes the basic processing calculation of an artificial neuron. 

There are several components that are valid for all artificial neurons, in spite of      their
position in the whole network (whether input, output or hidden layer); weights,      summation
functions, activation functions, transfer function, error target function \cite{Haykin}.   The
assembling of the neurons in layers, the provision of connections between the neurons   in the
layers, the summation and transfer function are common to all neural networks and    represent
the base of a neural network. In our specific implementation, the neuron's activation function 
for $\sigma_{i}(x)$ used in the network is the function $tanh(x)$. This is a rescaled  sigmoid
functions with output range $[-1,1]$.

\bigskip

\begin{figure}[h!]
\centering
\begin{minipage}{0.85\textwidth}
\begin{tabular}{c}
\includegraphics[width=1.0\textwidth]{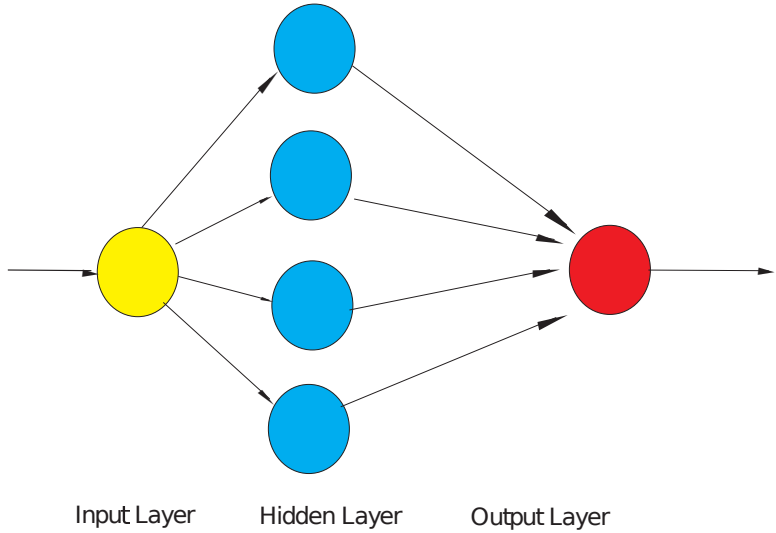}
\end{tabular}
\end{minipage}
\caption{Schematic of the architecture of the neural network implemented in this work.
The dots represent neurons, the lines express the connections between the neurons. The input layer
consist of one neuron, the hidden layer is implemented by four neurons, and the output layer
has one neuron.}
\label{fig:NNarchitecture}
\end{figure}

\label{Learning Strategy}
Many types of learning strategies exist, usually determined by the way the values change, which
main goal is to reach a good balance between finding a sufficiently accurate (usually    local)
minimum and computational resources \cite{Haykin}. Most widely used learning strategies are the
Hebb's rule \cite{Herz}, the Hopfield law \cite{Hopfield}, and the delta rule (also known    as
the Least Mean Square Learning Rule). In this particular work we have developed a   three-layer
feedforward network (see Fig.~\ref{fig:NNarchitecture}), consisting of one input neuron,    $4$
hidden neurons and one output neuron capable of approximating a non-linear function   (although
this is certainly not the only possible choice). As previously mentioned the number of   hidden
neurons is critical to the network performance. During the design of the network \cite{Bishop},
\cite{Fujita}, \cite{Hornik}, we considered the inner layer neuron count, accounting for    the
fact that more neurons will provide the capability to approximate functions of great complexity.
At the same time, populating the hidden layer with too many neurons could lead to   overfitting
the training data and thus generate worse results. On the other hand, too few nodes      in the
hidden layer will lead to lack of power to provide desired outputs.

The training process is based on the simulated annealing method, which consists of two   steps:
first of all, an effective temperature is increased to a maximum value, secondly the  effective
temperature is decreased slowly until the particles (representing the solution of           the
optimization problem) rearrange themselves in the ground state of the solid. Eventually     the
probability of a move for a point is given by
\begin{equation}
 Pr[accept]=e^{\frac{-\Delta E}{T}},
\end{equation}
where $\Delta E$ is the difference between the actual energy and the energy before the    move,
and $T$ is the effective temperature of the system \cite{Kirkpatrick}. Therefore a probabilistic
acceptance is achieved by generating a random number $R$ in the range $[0,1]$, which is    then
compared to $Pr[accept]$ and if $R<Pr[accept]$, the move is accepted. When the     optimization
problem achieves a lower effective temperature, fewer instances are accepted of          larger
temperatures and it resembles closely downhill-only improvement.

\subsection{Miming quantum randomness}
\label{mimick-quantum-randomness}

The possibility of biological neural network exploiting quantum effects strongly suggests   the
opportunity of introducing randomness in a neural network which could, somehow,       introduce
computational advantages \cite{consciousness}. At a first glance, in the context of ANN,   this
may suggest the use of quantum mechanical laws inside the very core of a neuron. In   practice,
this would correspond to the necessity of numerically simulating the             time-dependent
Schr{\"o}dinger's equation (or any other equivalent formalism such as Feynman, Wigner, etc)  to
quantitatively determine the eventual tunnelling effects. This would amount to      numerically
simulate the following time-dependent partial differential equation \cite{Goldberg}
\begin{eqnarray}
i \hbar \frac{\partial \Phi}{\partial t} \left( {\bf{r}}, t \right) =
 \left( -\frac{\hbar^2 \nabla^2}{2 m} + V \left( {\bf{r}} \right) \right) \Phi \left( {\bf{r}}, t \right),
\end{eqnarray}
where $i$ is the imaginary unit, $\Phi ({\bf{r}}, t)$ is the wave function defined over   space
and time, $\hbar$ is the reduced Planck's constant, $r$ is the position of the particle, $t$ is
the time, $m$ is the mass of the particle, $\nabla^2$ is the Laplacian operator,            and
$V({\bf{r}},t)$ is the potential energy acting on the particle. While this task would        be
definitely affordable for a relatively small number of (independent) artificial neurons,     it
represent a daunting task in the context of ANN where one may have to deal with thousands    of
neurons.
Therefore we suggest a computationally more convenient technique which aim is the miming of the
presence of randomness, intrinsic of a quantum system, without the burdening of            high
computational costs related to highly accurate quantum simulations. 

\bigskip

\begin{figure}[h!]
\centering
\begin{minipage}{0.9\textwidth}
\begin{tabular}{c}
\includegraphics[width=0.55\textwidth]{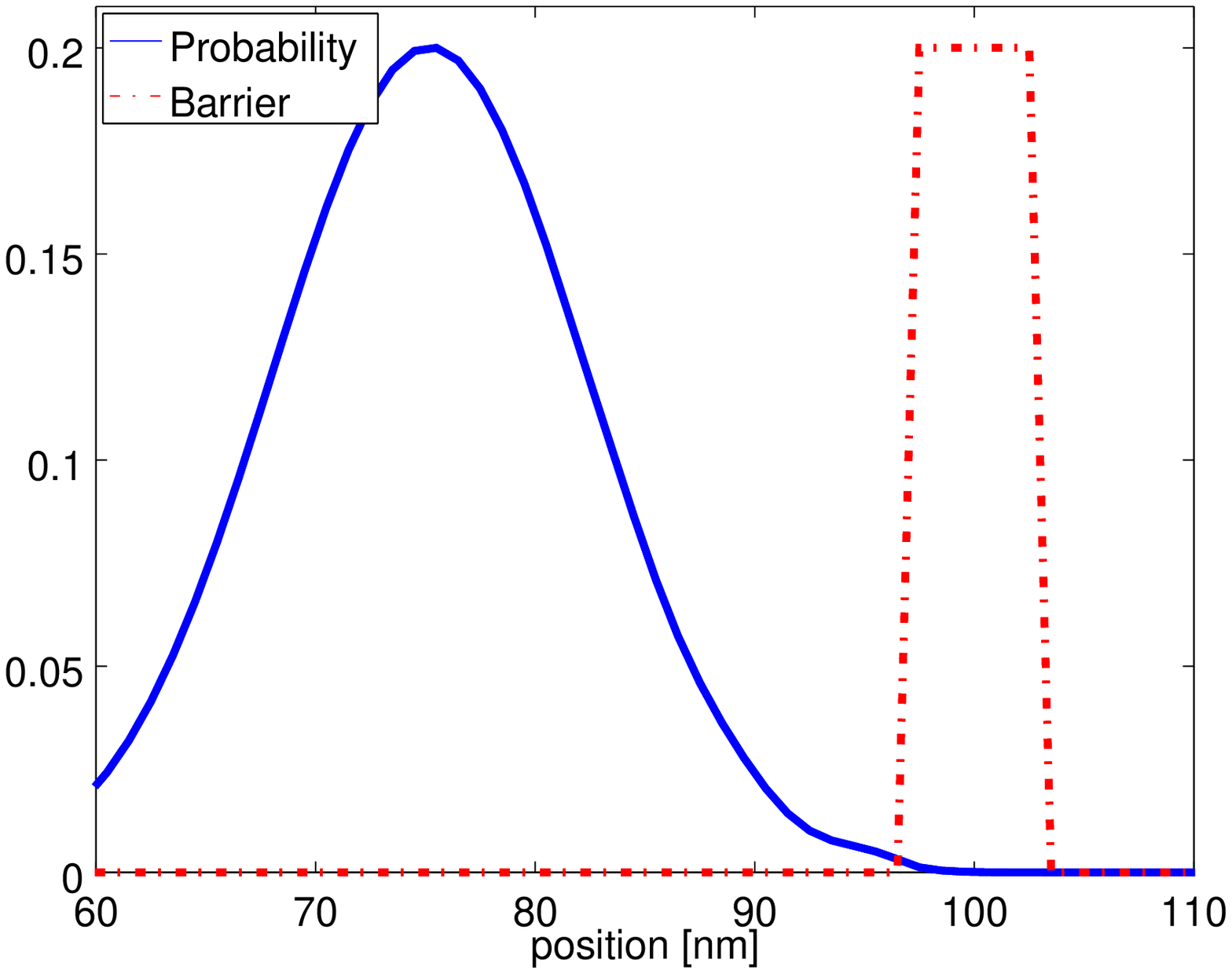}
\includegraphics[width=0.55\textwidth]{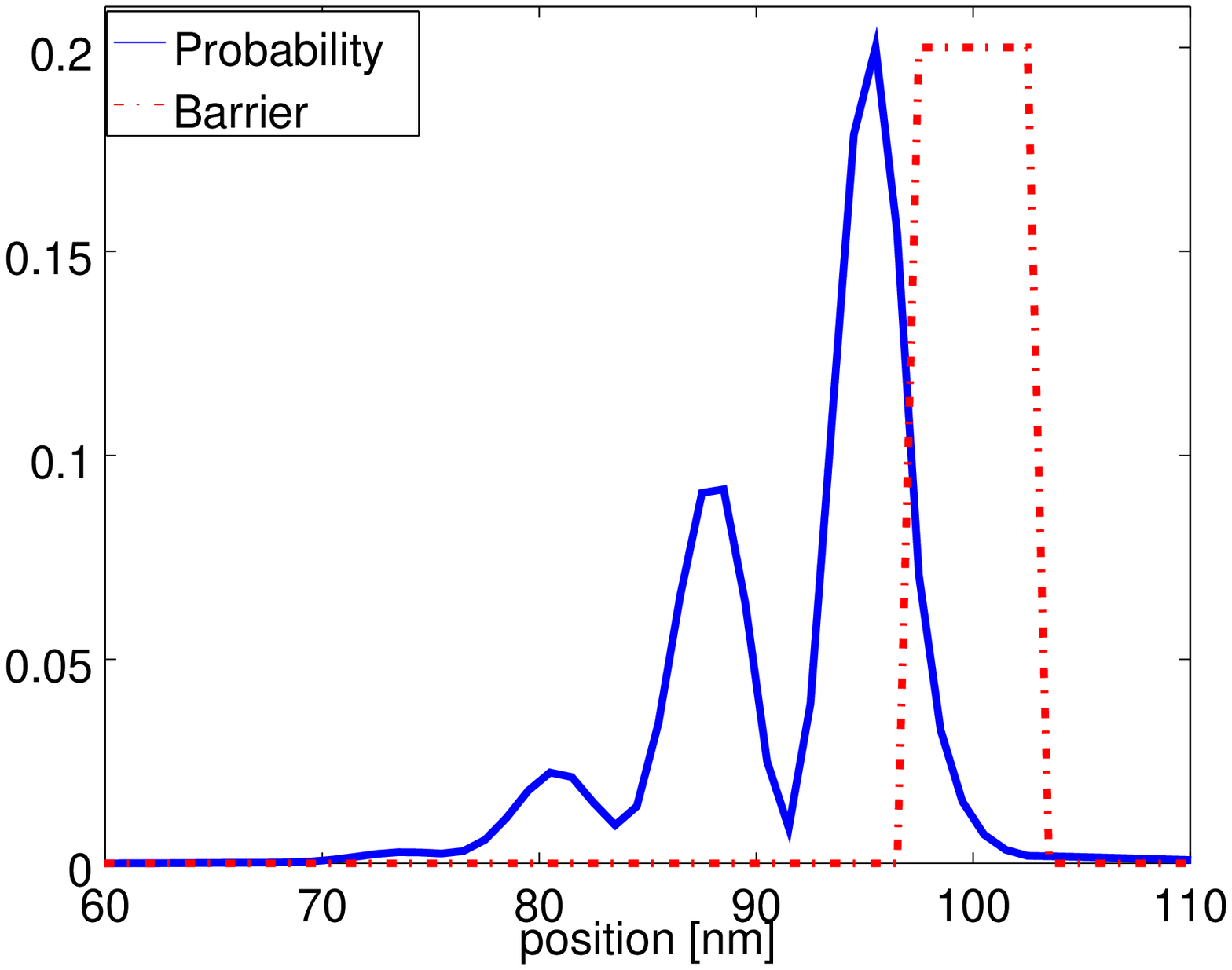}
\end{tabular}
\end{minipage}
\caption{Left plot: a Gaussian wave-packet, (blue) continuous line, is travelling against
an energetic potential barrier, (red) dashed line. Right plot: after a certain time, the wave-packet is
interacting with the barrier. Part of the packet is scattering back while the rest is tunnelling.}
\label{fig:potential_barrier}
\end{figure}

The technique is based on the following fact: it is possible to model, by analogy, a biological
neuron as a semiconductor heterostructure consisting of one energetic barrier     (e.g. AlGaAs)
sandwiched between two energetically lower areas (e.g. GaAs) \cite{consciousness}.   Therefore,
the activation function of an artificial neuron can be viewed as one or more particles entering
the heterostructure and interacting with the barrier    (see Fig.~\ref{fig:potential_barrier}).
The modulus of its wave-function $|\Phi ({\bf{r}}, t)|^2$ provides the probability of   finding
the particle in some point of the device at time $t$, thus introducing randomness        in the
process (Born rule). If the probability of back scattering is higher than the probability    of
tunnelling we consider the activation function inhibited, and vice versa.

The reader might establish similarities between our idea and the main structure of an    Action
Potential, which could also be described as a distinct voltage-gated ion channels in   a cell's
membrane. Once the potential increases to a defined threshold value, the membrane potential  of
the cell opens. The membrane potential maintains an electric potential difference    (voltage),
which once triggered, is activated.

\bigskip

In practice, we achieve this goal by adding in the network  a function \textbf{\textit{addHiddenNoise}}
which, for every neuron in the hidden layer, adds a noise to the already computed set of weights:
\begin{equation}
hiddenLayer.neurons[i].weights[j]+= randomDouble() * weightNoise
\end{equation}
which mathematically corresponds to the expression
\begin{equation}
(w_{1}x_{1}+w_{2}x_{2}+w_{3}x_{3}...+w_{n}x_{n}).
\end{equation}
In the context of ANN training, our suggested technique is in some sense comparable to certain
elements in the Genetic Algorithm (GA), specifically the  mutation part of the      algorithm.
The mutation step can be viewed as an initialization of random walks through the search  space
of possible solutions. The mutations are ordinarily small and are defined by step and     rate,
which can be constant or adaptive. The genetic algorithm involves the creation of          new
generations of individuals, from which the algorithm selects the best ones and further evolves
the population according to a predefined set of rules. During the training process, the   main
purpose of the mutation operator is to maintain the diversity within a population of       ANN
weights in order to prevent premature convergence.  Our technique, on the other hand        is
designed to reach a random improvement at very low computational cost. The procedure  involves
the generation of random doubles by means of a Mersenne Twister. The main purpose   remains to
restrict the noise inside a certain range, specified by the user. This is achieved by comparing
the new (random) set of weights to the one obtained by the classical network. Eventually   the
function \textbf{\textit{backupState}} is used to copy the weights for restoration when    the
addition of noise produces a larger error (root mean square) compared to the noiseless set  of
weights. In these terms, our novel technique provides an algorithm that incurs      negligible
computational cost since it depends on the simplicity of generating random numbers. 

\bigskip

In the next section, we describe several numerical experiments which aim is the validation  of
our approach.

\section{Numerical Validation}
\label{Numerical Validation}
In this section, we present a numerical validation of our suggested post-learning strategy for
a neural network. The network's aim is to fit two known functions - a polynomial of     second
degree ($f(x)=x^2$) and the square root of a polynomial ($f(x)=\sqrt{x}$), given three    data
points. The network architecture  is identical for both functions - three layers, with     one
input neuron, $4$ hidden neurons and one output neuron. Additionally, we have         assigned
limitations to the weights space. In the first case, the weights search is constrained between
$-12 (min)$ and $+12 (max)$. The weights for the second function are randomly      distributed
between $-1 (min)$ and $+1 (max)$. The training process is based on the simulated    annealing
method for either function.

In the current situation we deliberately stop the network at an arbitrary local minimum     by
means of a temperature rate, decreasing in a non-optimal fashion. This is done in order     to
clearly show that our technique can provide a way to further improve the training even   after
the optimization process. This situation is of importance as with nowadays available big  data
it becomes difficult to find the best set of weights for a ANN due the ever growing complexity.
Learning is performed by utilizing three equidistant data points, belonging to the       range
$[0,1]$, excluding the extrema. The first case (see Fig.~\ref{fig:squareoutput}, upper    left
side) utilizes the following data points $(0.1, 0.01)$, $(0.5, 0.25)$ and $(0.9, 0.81)$.   The
three data points, exploited for training the network for square root of polynomial are     as
follows: $(0.15, 0.38)$, $(0.6, 0.77)$, $(0.85, 0.92)$. 

To understand the extent of influence of noise on the network's output and error decrease,  we
investigated $5$ different scenarios. First we run the network without noise addition in order
to benchmark the quantum part of the network to the classical                              one
(see Fig.~\ref{fig:squareoutput} and Fig.~\ref{fig:sqrteoutput} upper left plots). The rest of
the tests include the network working with $0.5\%$, $1\%$, $2\%$, and $4\%$ noise respectively. 

The following two subsections attempt to explain how the level of noise is affecting the   two
functions. 
\subsection{Polynomial of second degree}
\label{polynomial2nddegree}
\begin{figure}[h!]
\centering
\begin{minipage}{1.0\textwidth}
\begin{tabular}{c}
\includegraphics[width=0.55\textwidth]{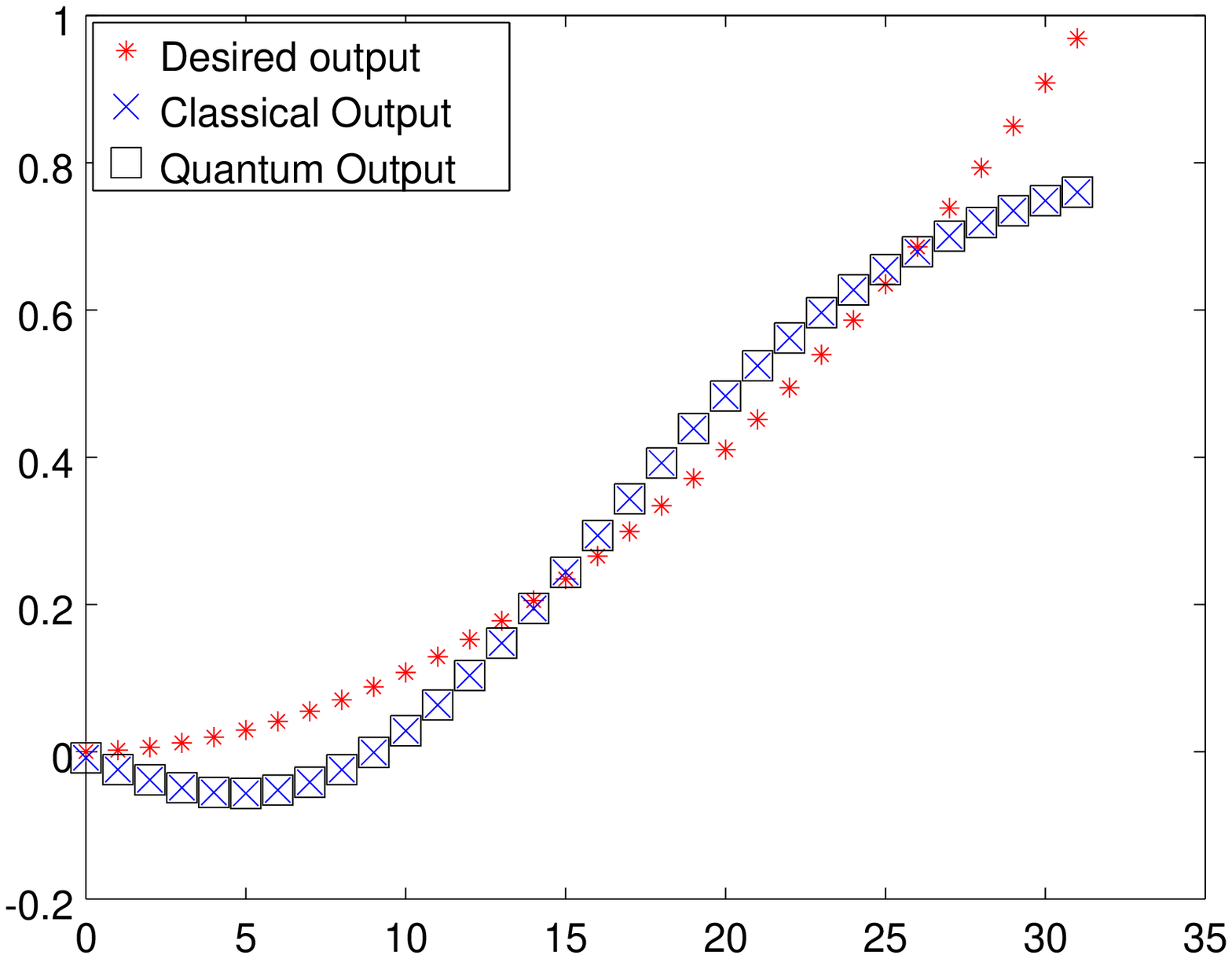}
\includegraphics[width=0.55\textwidth]{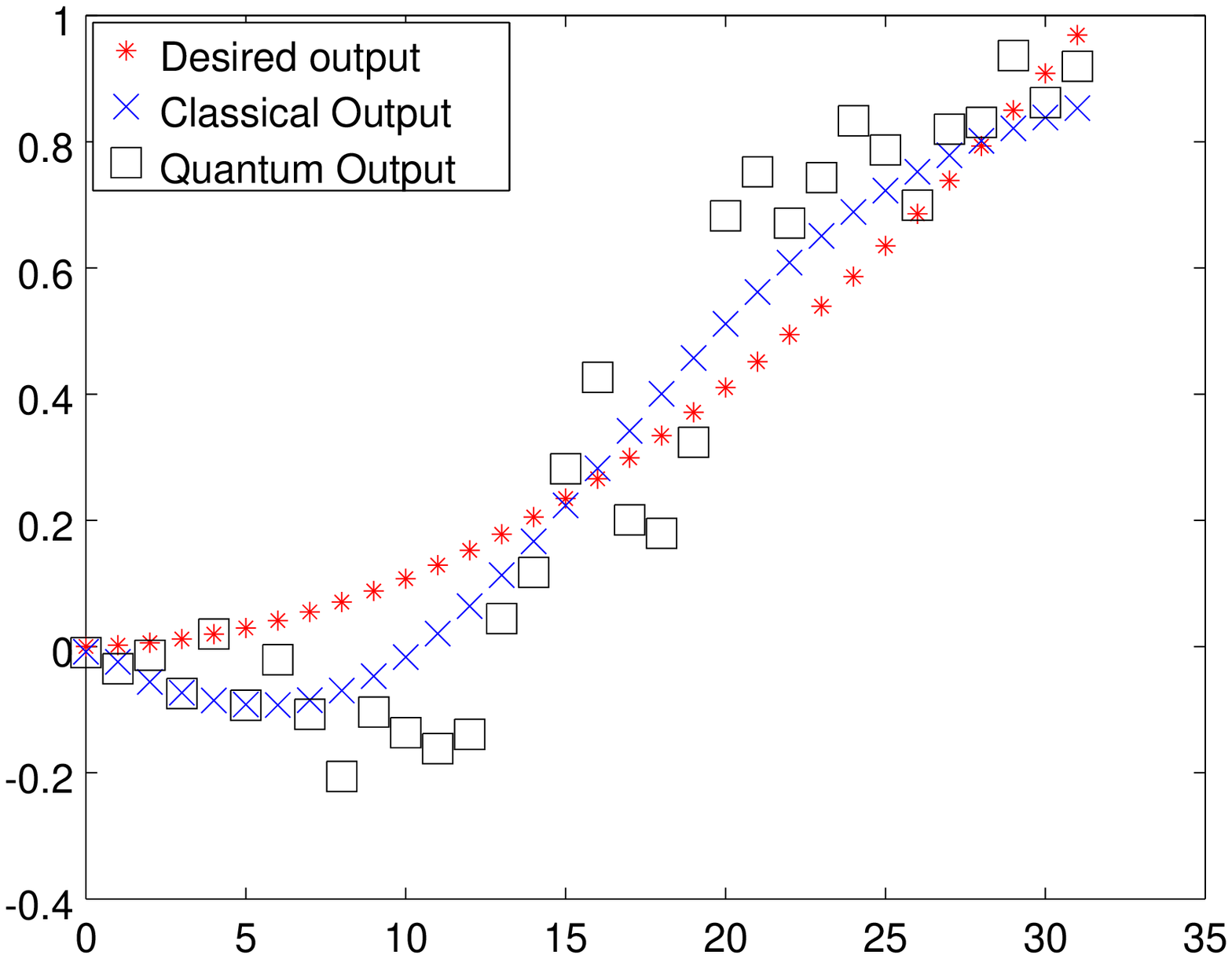}
\\
\includegraphics[width=0.55\textwidth]{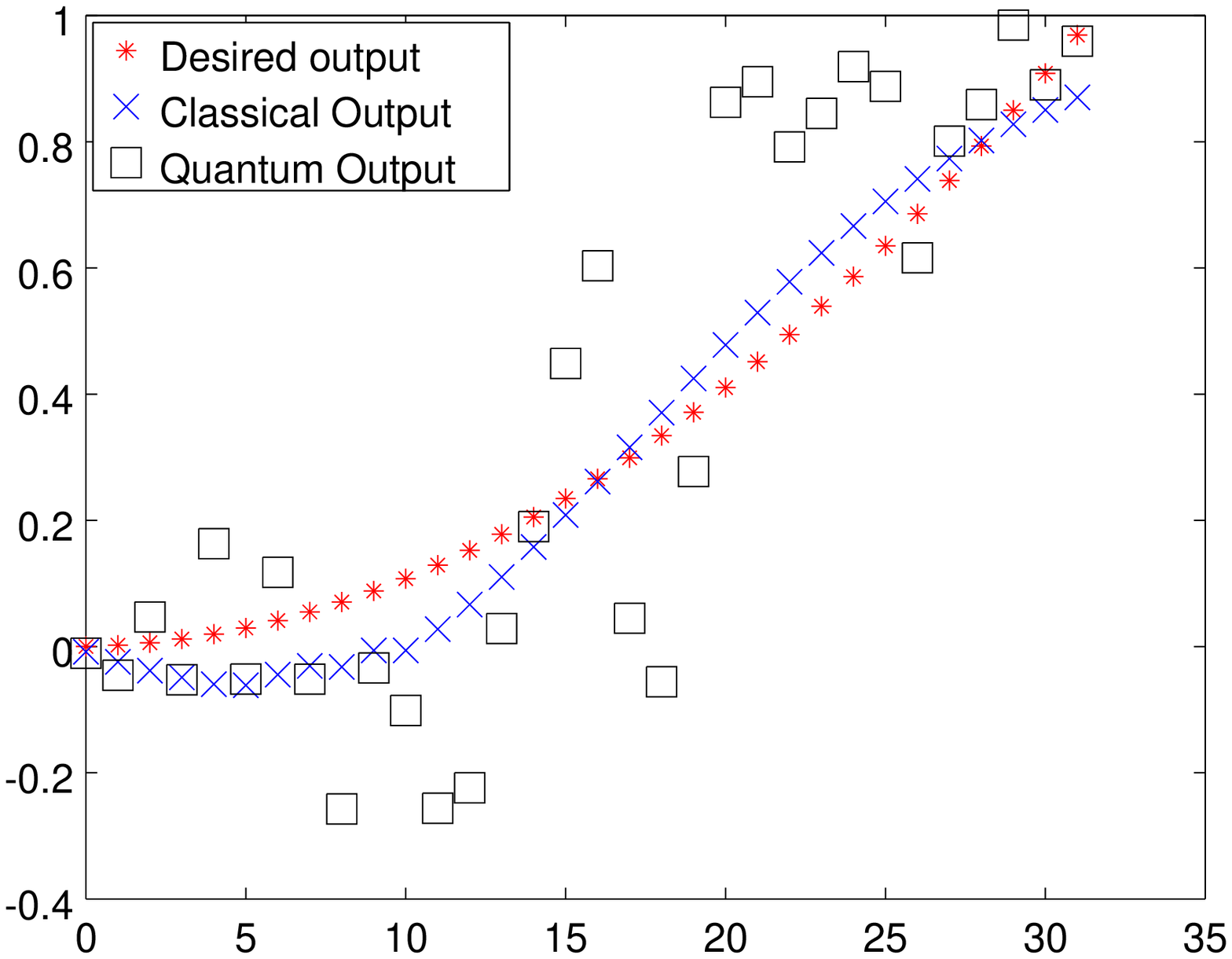}
\includegraphics[width=0.55\textwidth]{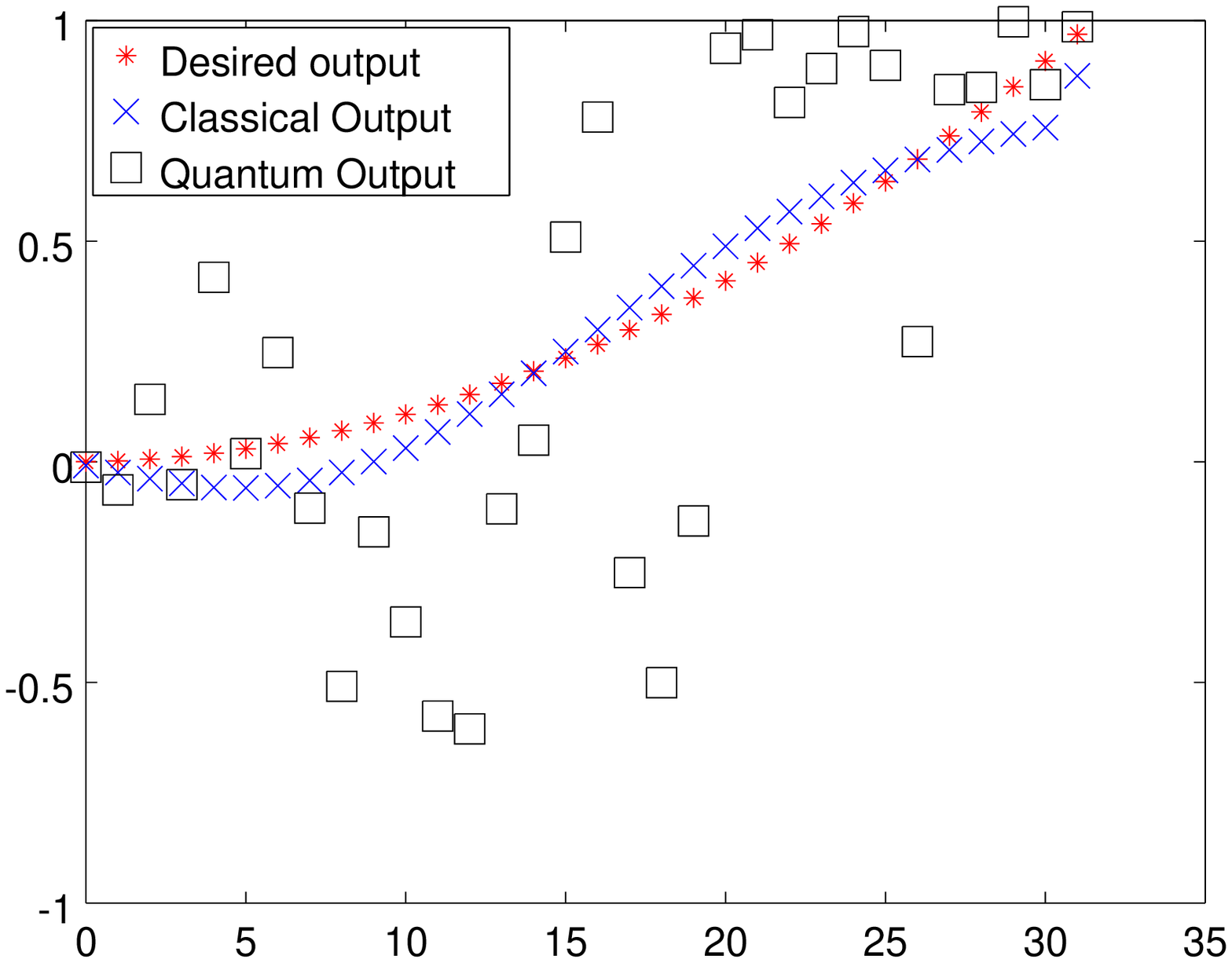}
\\
\includegraphics[width=0.55\textwidth]{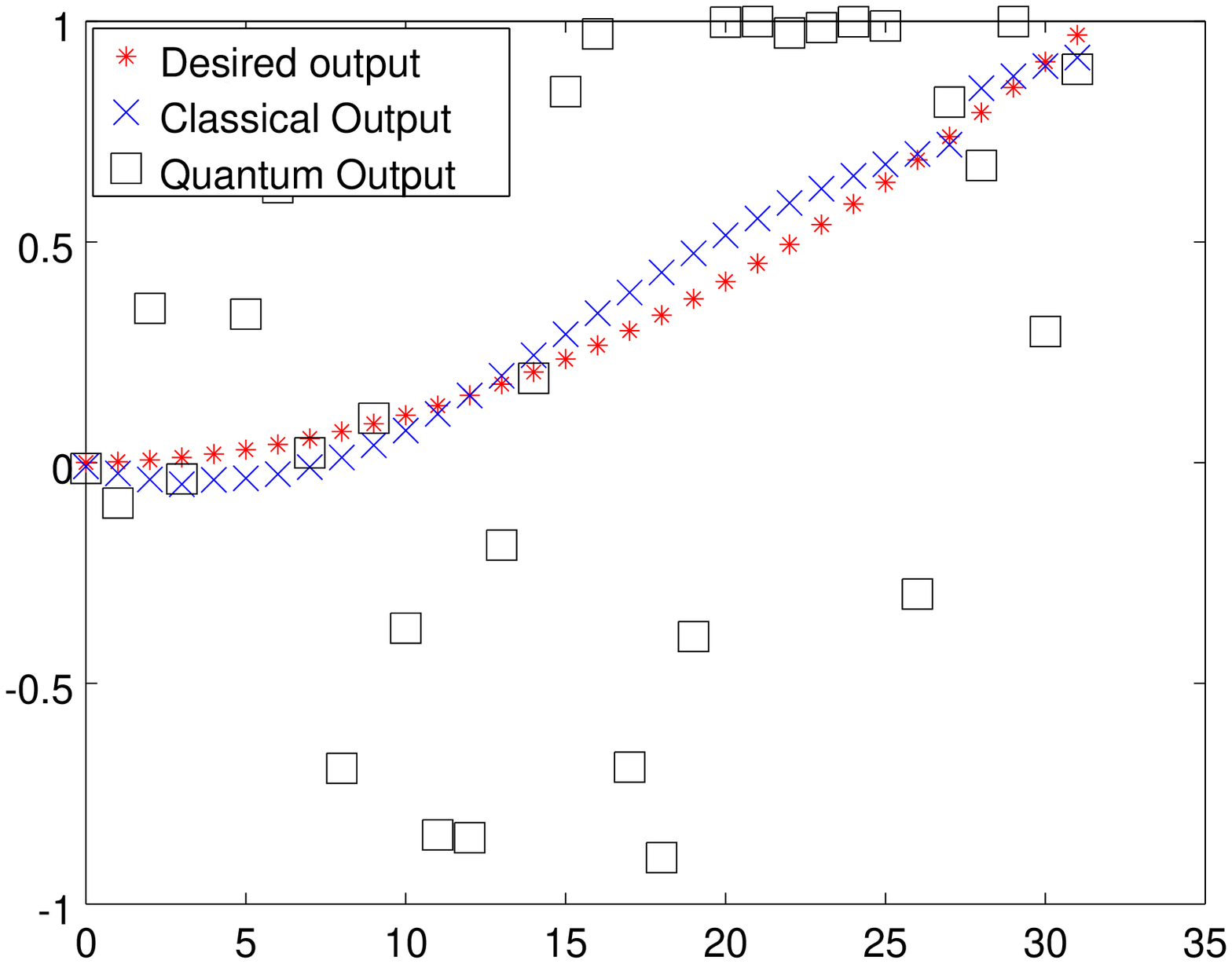}
\end{tabular}
\end{minipage}
\caption{The plots feature the output of our neural network for the function of polynomial of second degree. 
The (red) star symbolizes the desired network's output, the (blue) $x$ indicates the output from the classical mode, 
and the quantum mode is denoted by a square. The upper left side plot exhibits a validation test 
for the network, running in both classical and quantum mode, with no noise applied. The upper right 
plot represents the output when $0.5\%$ noise is applied. The middle left and right plot, display 
the network's output when $1\%$ and $2\%$ noise is assigned respectively. 
The final plots consists of the output when the network is executed with $4\%$ noise.}
\label{fig:squareoutput}
\end{figure}

\begin{figure}[h!]
\centering
\begin{minipage}{1.0\textwidth}
\begin{tabular}{c}
\includegraphics[width=0.55\textwidth]{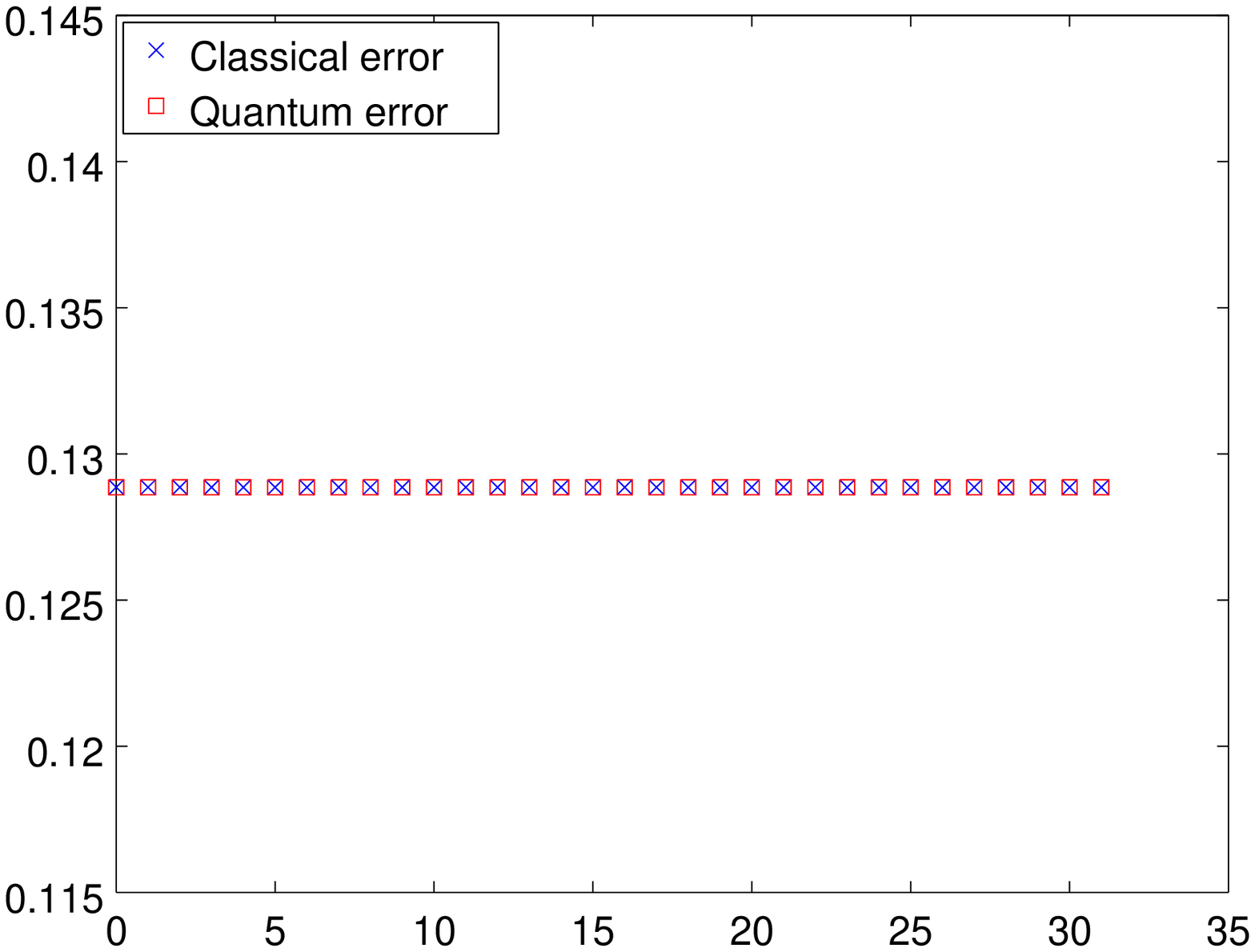}
\includegraphics[width=0.55\textwidth]{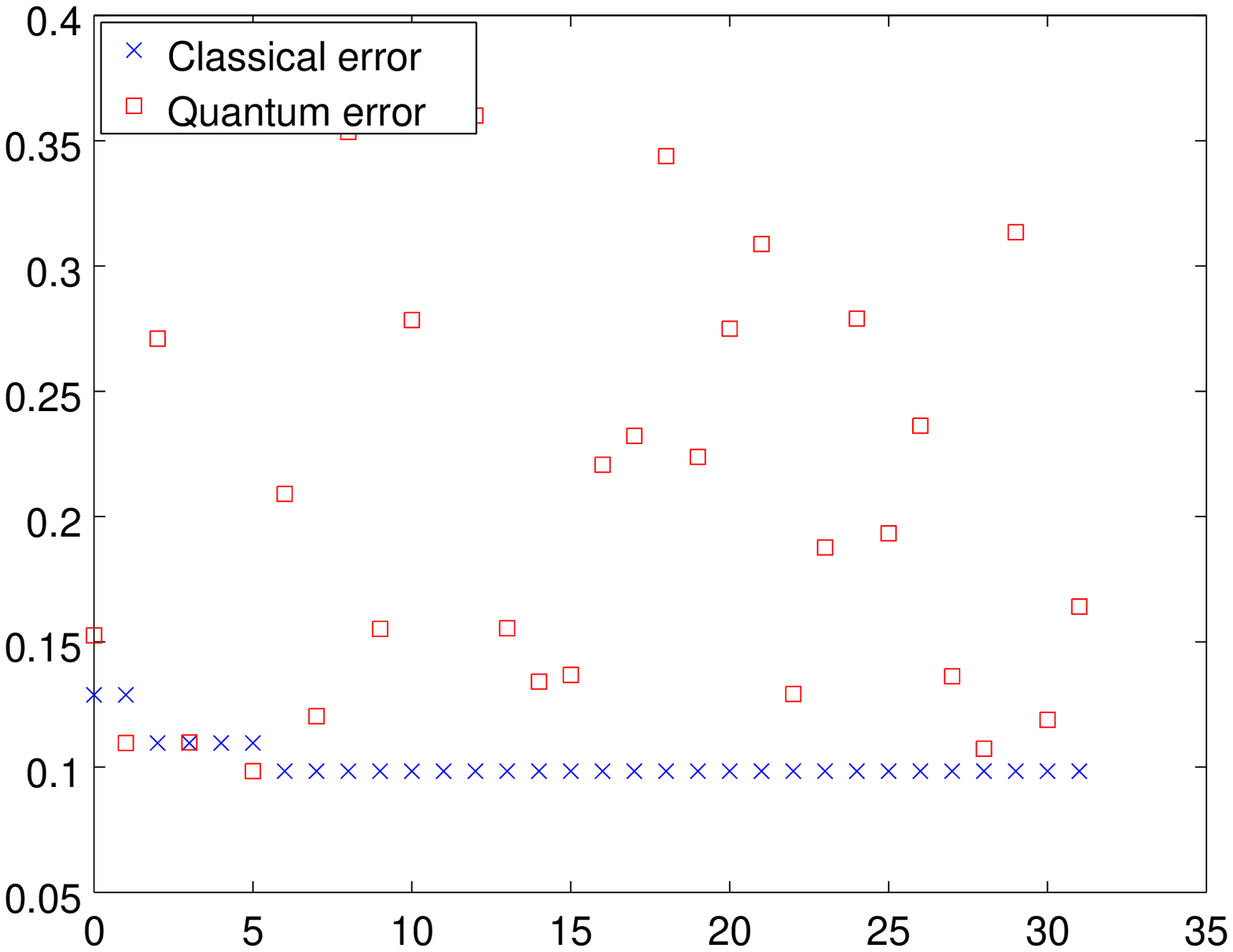}
\\
\includegraphics[width=0.55\textwidth]{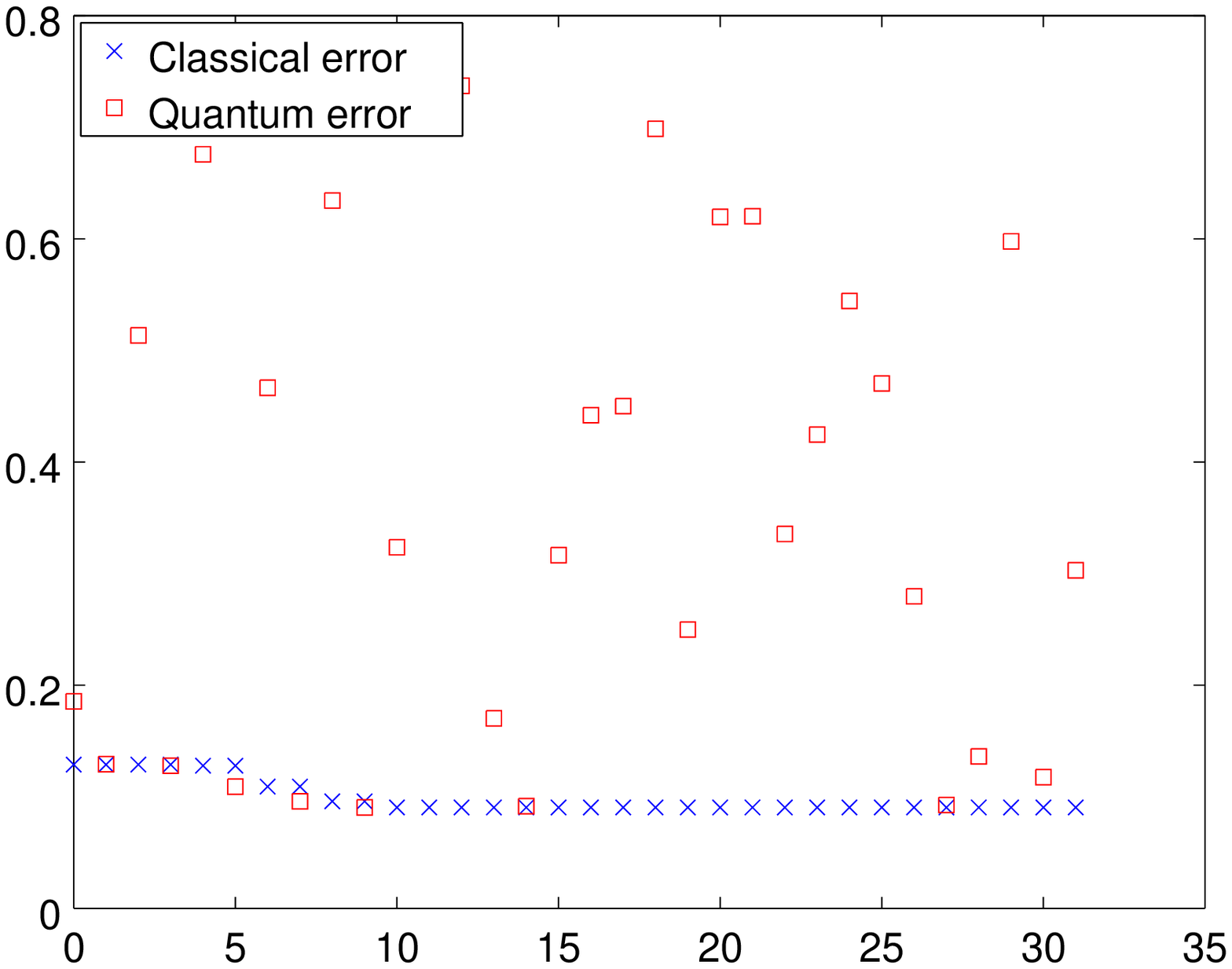}
\includegraphics[width=0.55\textwidth]{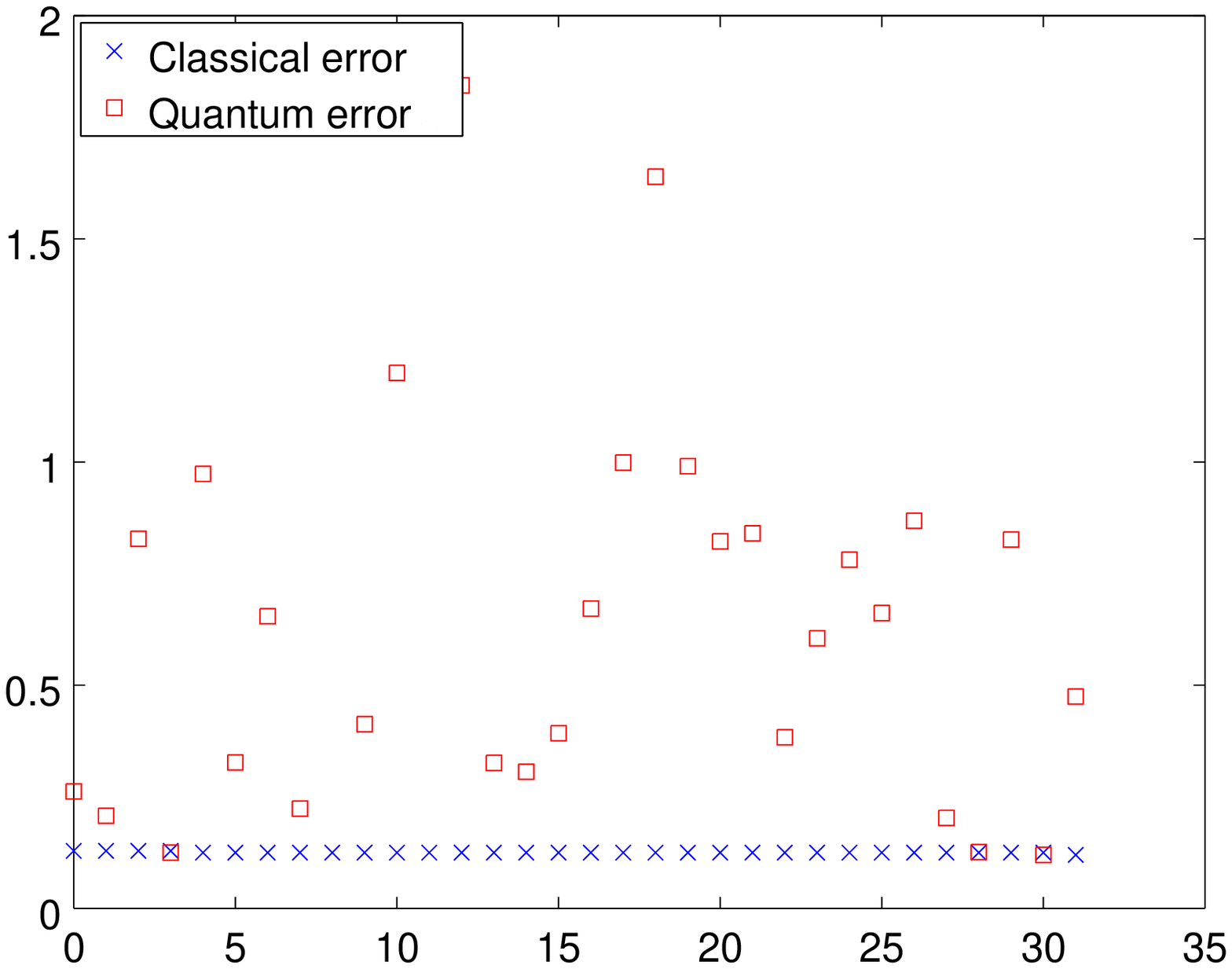}
\\
\includegraphics[width=0.55\textwidth]{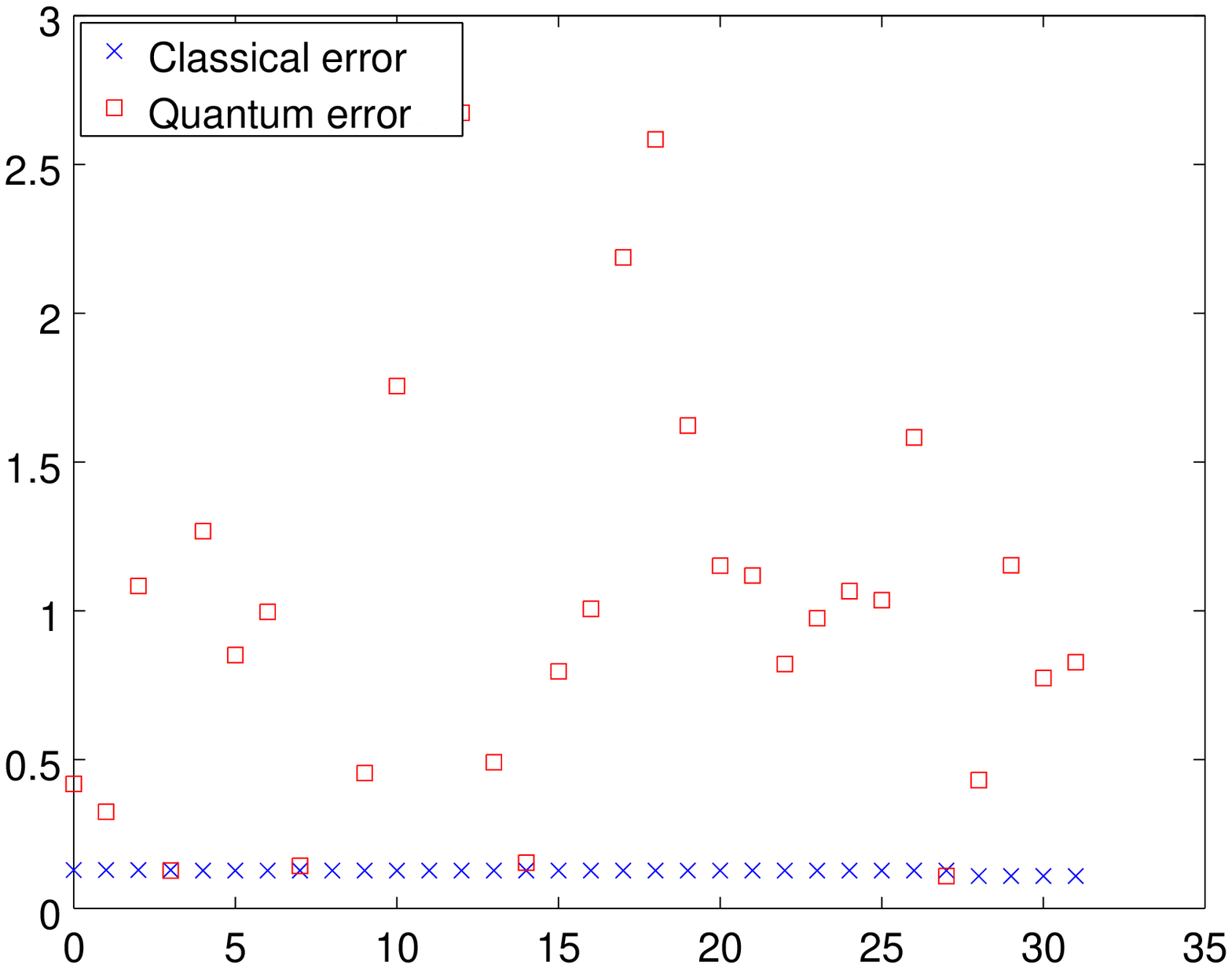}
\end{tabular}
\end{minipage}
\caption{Error reduction from the neural network in the case of polynomial of second degree. The error from the 
classical part of the network is depicted as a (blue) $x$, while the error from the quantum side of the network is
shown as a (red) square. The upper left plot illustrates the network running with no noise for validation purposes.
Upper right plot illustrates the network's error in scenario with $0.5\%$ applied. The middle left and right plots
constitute network error in simulations with $1\%$ and $2\%$ noise. The left lowest plot represents error with $4\%$ 
noise.}  
\label{fig:errorsquareoutput}
\end{figure}

Our first validation experiment addresses the fitting of a polynomial of second degree.     To
ascertain the network performance we initially run the network without any noise to provide  a
benchmark test for its correct functioning. 
To observe how the levels of noise affect the network's output, we have initialized a     case
where we add minimal noise to the system ($0.5\%$). The effect on the quantum comportment   of
the network are minimal (see Fig.~\ref{fig:squareoutput} upper right plot), influencing    the
first few outputs. Corroboration is available from the error of the network (as seen        in
Fig.~\ref{fig:errorsquareoutput} upper right plot). As expected, for the first few data points,
the quantum part of the network contributes to slight improvement of the result. To    achieve
this, at every point, the algorithm compares the classical to the quantum error, choosing  the
better option, and discarding the other. Following a better quantum solution, the      network
accepts it and continues forward.  

Indicative to the amount of noise applied is the dispersement of quantum output and      error
through the plot in the situations where the amplitude of the noise is successively increased.
The moderate enlargement of noise to $1\%$ reveals that the network output in the middle   of
the solutions is less accurate, with better results in the upper and lower bound of the curve.

The network precision further deteriorates when we apply $2\%$ noise. This case supplies only
few reliable outputs close to the fitting curve. One could note that the increase in    noise
contributes to more output outliers in the outcome      (shown in Fig.~\ref{fig:squareoutput}
middle right plot). However, with addition of noise, the quantum network outperforms      the
classical one in very few situations. Moving from $1\%$ to to $2\%$ and $4\%$ noise,      the
overall error tends to decrease slowly (Fig.~\ref{fig:errorsquareoutput}, middle right    and
lower left plots). 

In the current numerical validation, the performance of the network is stable with    $0.5\%$
and $1\%$ noise added. The quantum part produces smaller error from the beginning of      the
calculations, with a possibility to decrease the error almost instantaneously.

The higher levels of noise still accomplish small improvements to the network, but at     far
lesser scale and depth.
\subsection{Square root of a Polynomial}
\label{polynomialsqrt}
\begin{figure}[h!]
\centering
\begin{minipage}{1.0\textwidth}
\begin{tabular}{c}
\includegraphics[width=0.55\textwidth]{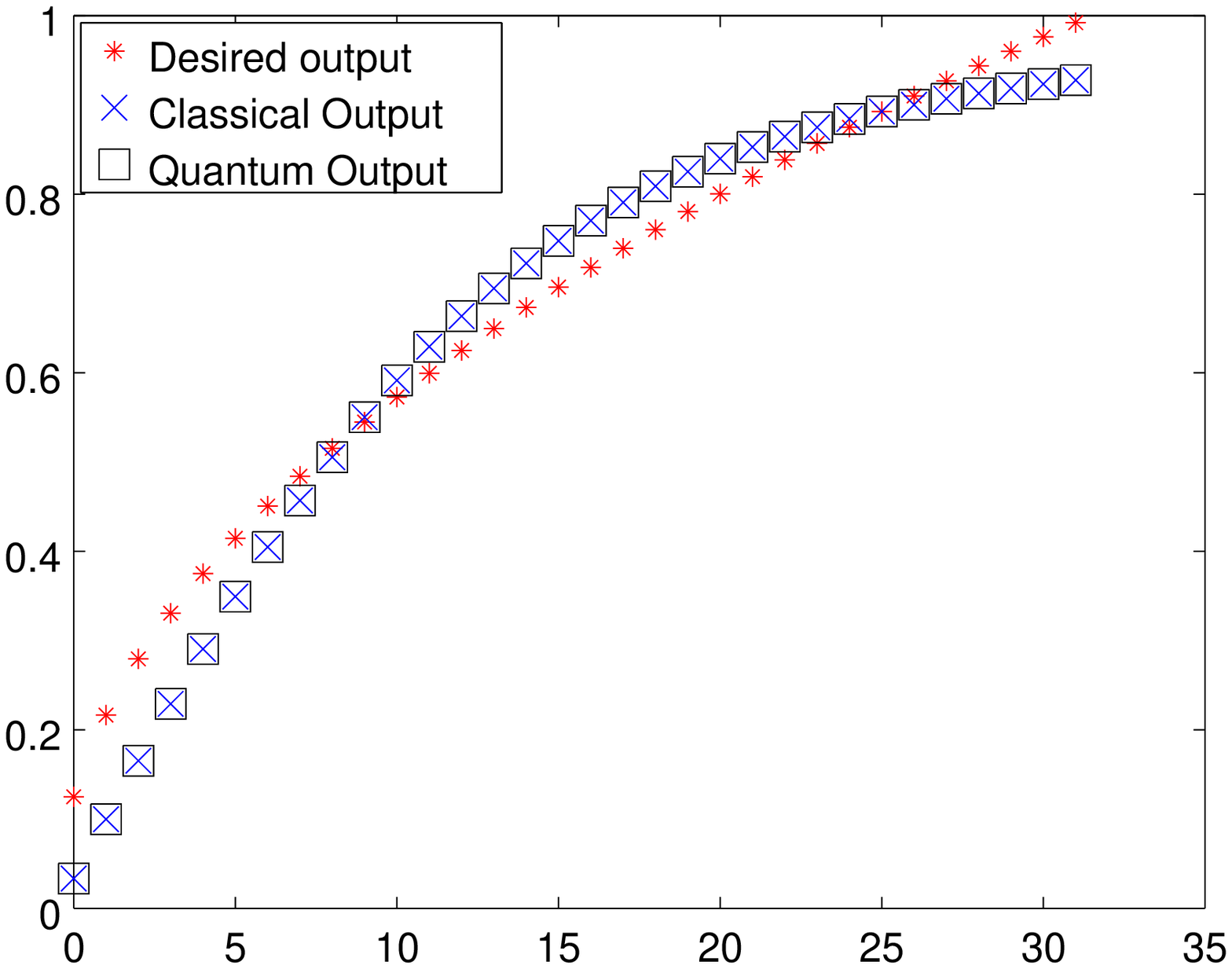}
\includegraphics[width=0.55\textwidth]{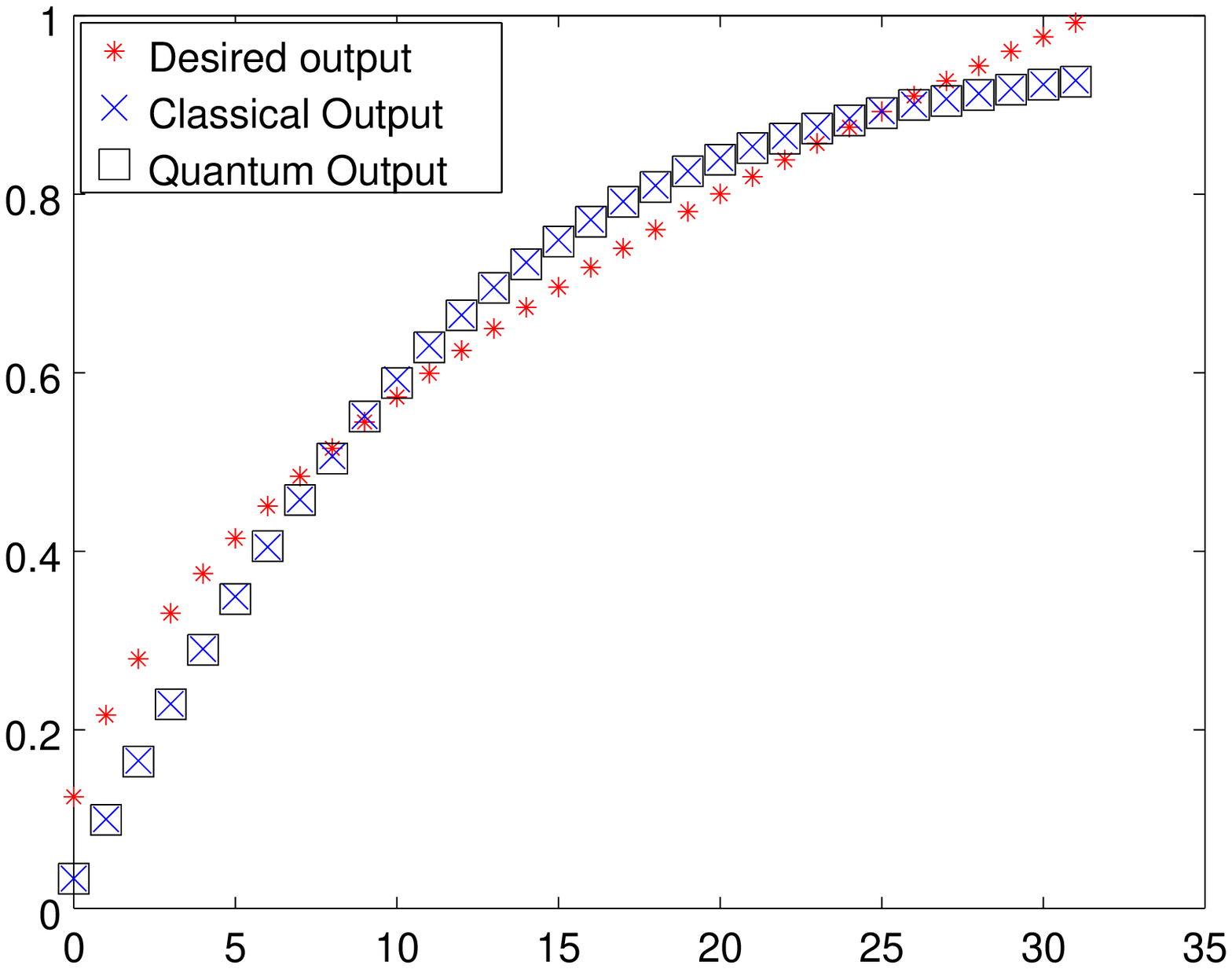}
\\
\includegraphics[width=0.55\textwidth]{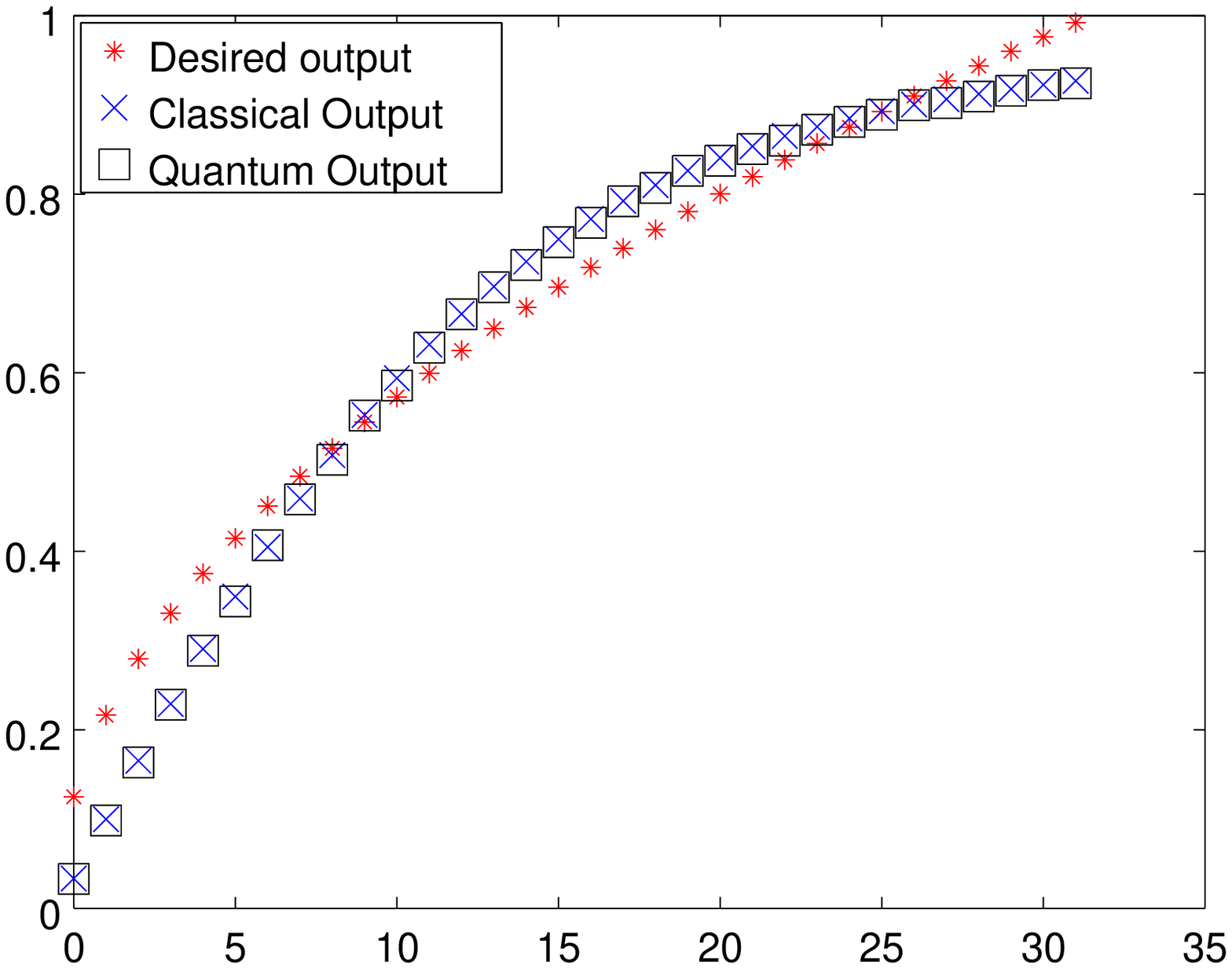}
\includegraphics[width=0.55\textwidth]{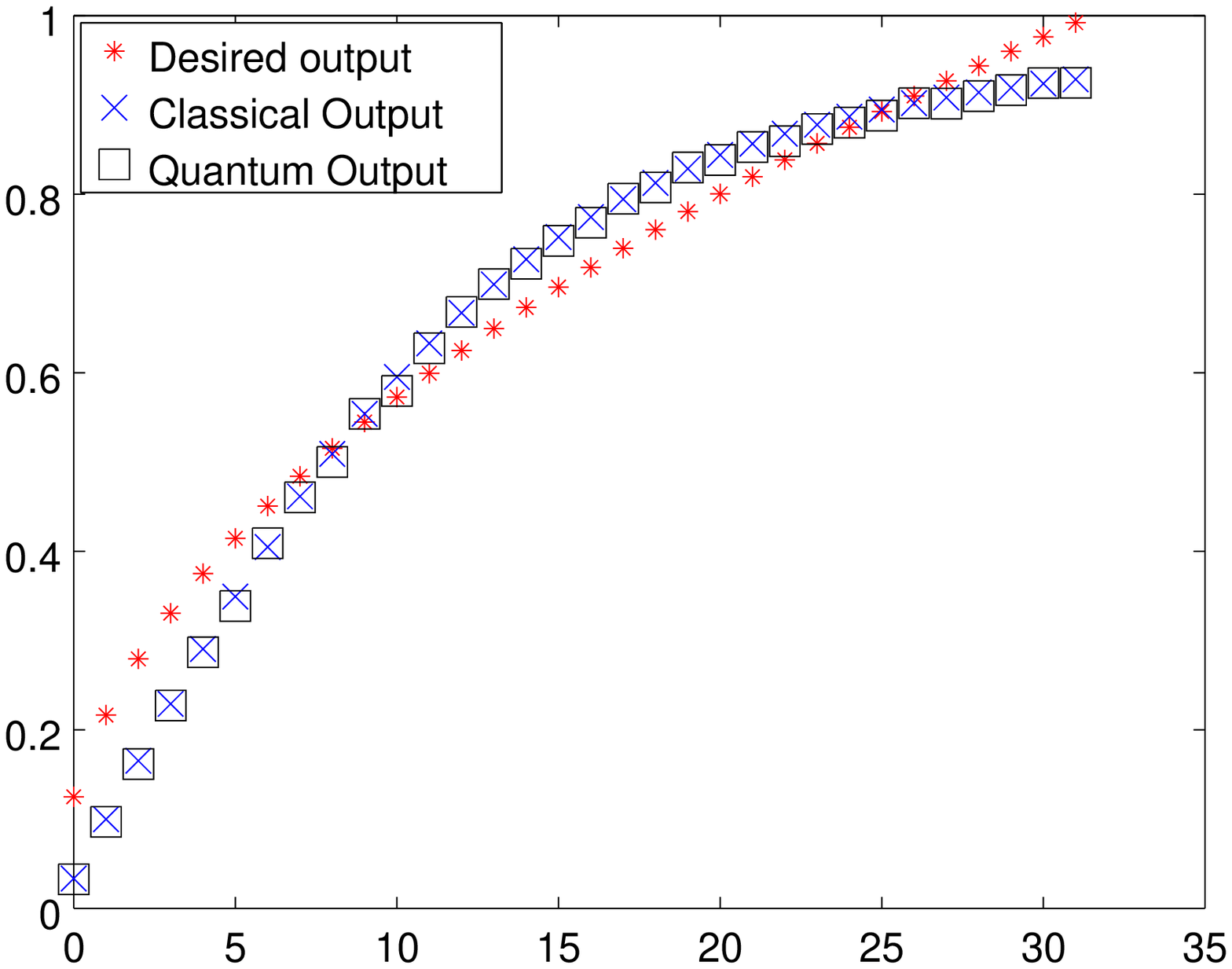}
\\
\includegraphics[width=0.55\textwidth]{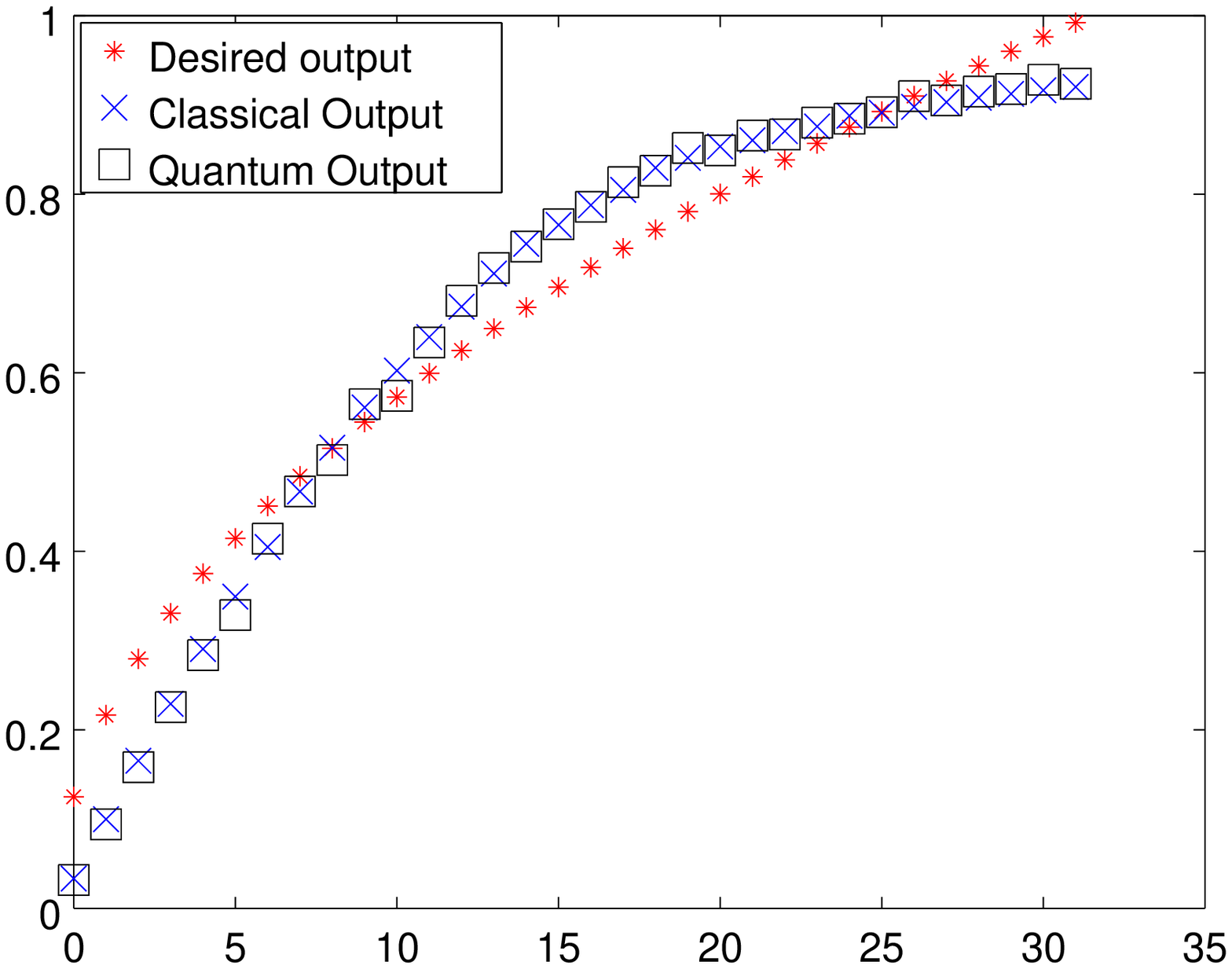}
\end{tabular}
\end{minipage}
\caption{Neural network outputs in the case of function of a square root of a polynomial. The (red) star symbolizes 
the desired network's output, the (blue) $x$ indicates the output from the classical mode, and the quantum mode is 
denoted by a square.The upper left and right plots represent network's output when $0\%$ and $0.5\%$ noise is executed. 
The middle left and right plots exhibit the output affected from $1\%$ and $2\%$ noise respectively. The lowest left plot
represent results from executing the network with $4\%$ noise.}
\label{fig:sqrteoutput}
\end{figure}

Considering the implication that our proposed technique could perform in a distinctive manner
for computing different functions, we have executed second test in fitting a square root   of
a polynomial. As the amount of noise was set to $0.5\%$, the results between the quantum part
resembled the classical ones (Fig.~\ref{fig:sqrteoutput} upper right plot). The pattern    of
error reduction is stepwise, appearing like the one in the similar situation for the    first
function. In this setup, only few better outcomes are provided, notwithstanding the     steep
descent of the error from point to point.

Our further analysis from the addition of $1\%$ noise, finds an outcome close to the  optimal
(Fig.~\ref{fig:sqrteoutput} middle left plot). In fact, the error continues to taper off in a
similar manner to the case from the $0.5\%$ noise. One would perceive, in accordance to   the
previous numerical validation test, that the increase in noise will decrease          network
performance. The current experiment exhibits the opposite direction.

The simulation shows that $2\%$ noise contributes to the increased efficiency of the  quantum
network. The network provides as much as twice as many quantum outcomes, as when     imposing
$0.5\%$ or $1\%$ noise, as well as lower errors (see Fig.~\ref{fig:errorsquareoutput}  middle
right plot). After the initial steep error reduction from the quantum part,       significant
divergence from the initial experiment is the clustering of the quantum error points near the
classical ones. 

In the next step, at $4\%$ noise, the error is consecutively decreasing, compared to any other
amount of noise administered. The reader should note that in this                    situation
(Fig.~\ref{fig:sqrteoutput} lower left plot), the network provides as much as twice as    many
quantum outcomes, as when only $0.5\%$ is applied. Throughout the various noise scenarios, the
outputs from the quantum network are analogue to the classical one. 
 
One possible explanation for the difference of network performance for the two       numerical
experiments could be provided by the variation of the weight space range for the two functions.
Further experiments are required to establish the network performance in relation to       the
weight space range, the noise amount and new functions. 

\begin{figure}[h!]
\centering
\begin{minipage}{1.0\textwidth}
\begin{tabular}{c}
\includegraphics[width=0.5\textwidth]{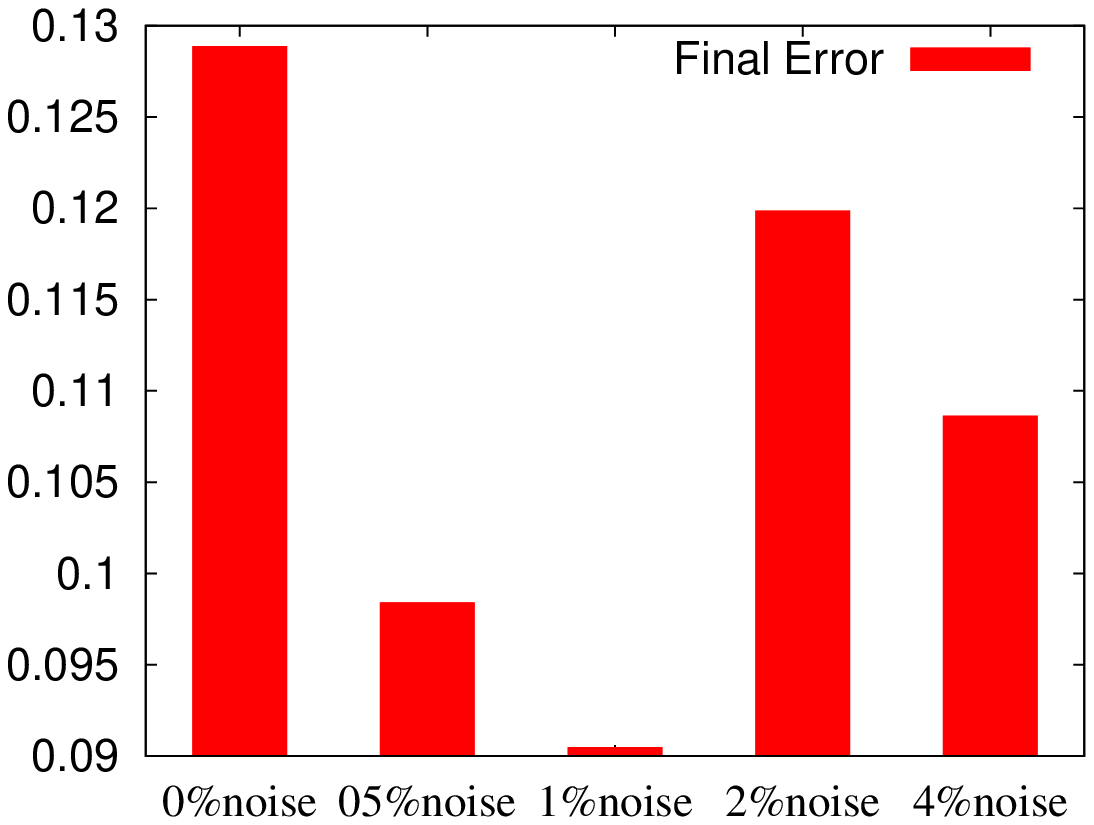}
\includegraphics[width=0.5\textwidth]{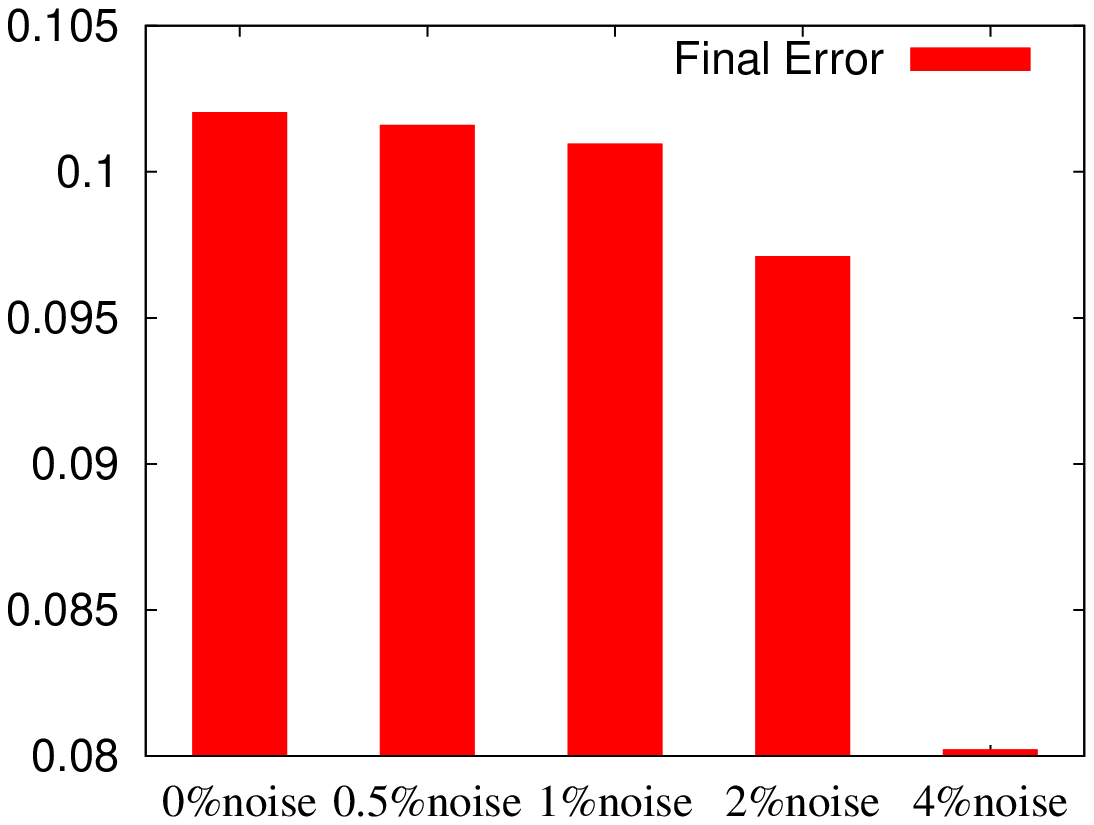}
\end{tabular}
\end{minipage}
\caption{Final error after the network calculates for $32$ output points.
The left side plot illustrates the error for the function of polynomial of second degree.
The right side plot depicts the error for the function of square root of 
a polynomial. The initial error for the left plot for all levels of noise is $0.12886$. The
initial error for the right plot for every noise level is $0.10203$.}
\label{fig:errorsquareroot}
\end{figure}

Illustration of the divergence of error reduction for the two known function is available from
Fig.~\ref{fig:errorsquareroot} (for the sake of clarity we also report these results in    the
shape of a table, see table~\ref{table:finalerror}). The left plot confirms the perception  of
increased network performance on noise levels at $0.5\%$ or $1\%$. The behavior of the network
is opposite for the second numerical experiment. Running the network with gradual increase  in
the noise actually contributes to the tapering off of the error.

\section{Conclusions}
\label{Conclusions}

In this article, we introduced a novel {\sl{post-learning strategy}} that is implemented as a
auxiliary reinforcement to the classical learning process of neural networks. The        main
purpose of this novel technique is to provide a method that is computationally reasonable  in
the scenario of a network trying to circumvent a local minimum during the training    process.
In order to achieve it, we suggested an approach based on the generation of random numbers at
the core of artificial neurons, which attempts to mimic the presence of quantum   randomness.
By performing several numerical experiments, we validated the method against the problem   of
fitting a known function, given a certain number of training points, and we have shown    how
our technique provides certain improvements in the system, without relevant        additional
computational costs. Certainly, further investigation is necessary to establish the     right
amount of noise, which has to be introduced in a network in order to achieve             real
improvements. This will be the subject of a future work.

\begin{table}
\caption{Final error estimation after the network calculates $32$ output points for
the two numerical validation functions. See also fig.~\ref{fig:errorsquareroot}	}
\label{table:finalerror}       
\begin{tabular}{lll}
\hline\noalign{\smallskip}
Noise & $f(x)=x^2$ & $f(x)=\sqrt(x)$  \\
\noalign{\smallskip}\hline\noalign{\smallskip}
0\%  &  0.1288600 & 0.1020300 \\
0.5\% & 0.0983995 & 0.1015950 \\
1.0\% & 0.0904621 & 0.1009560 \\
2.0\% & 0.1198510 & 0.0971008 \\
4.0\% & 0.1086160 & 0.0802115 \\
\noalign{\smallskip}\hline
\end{tabular}
\end{table}

\begin{figure}[h!]
\centering
\begin{minipage}{1.0\textwidth}
\begin{tabular}{c}
\includegraphics[width=0.55\textwidth]{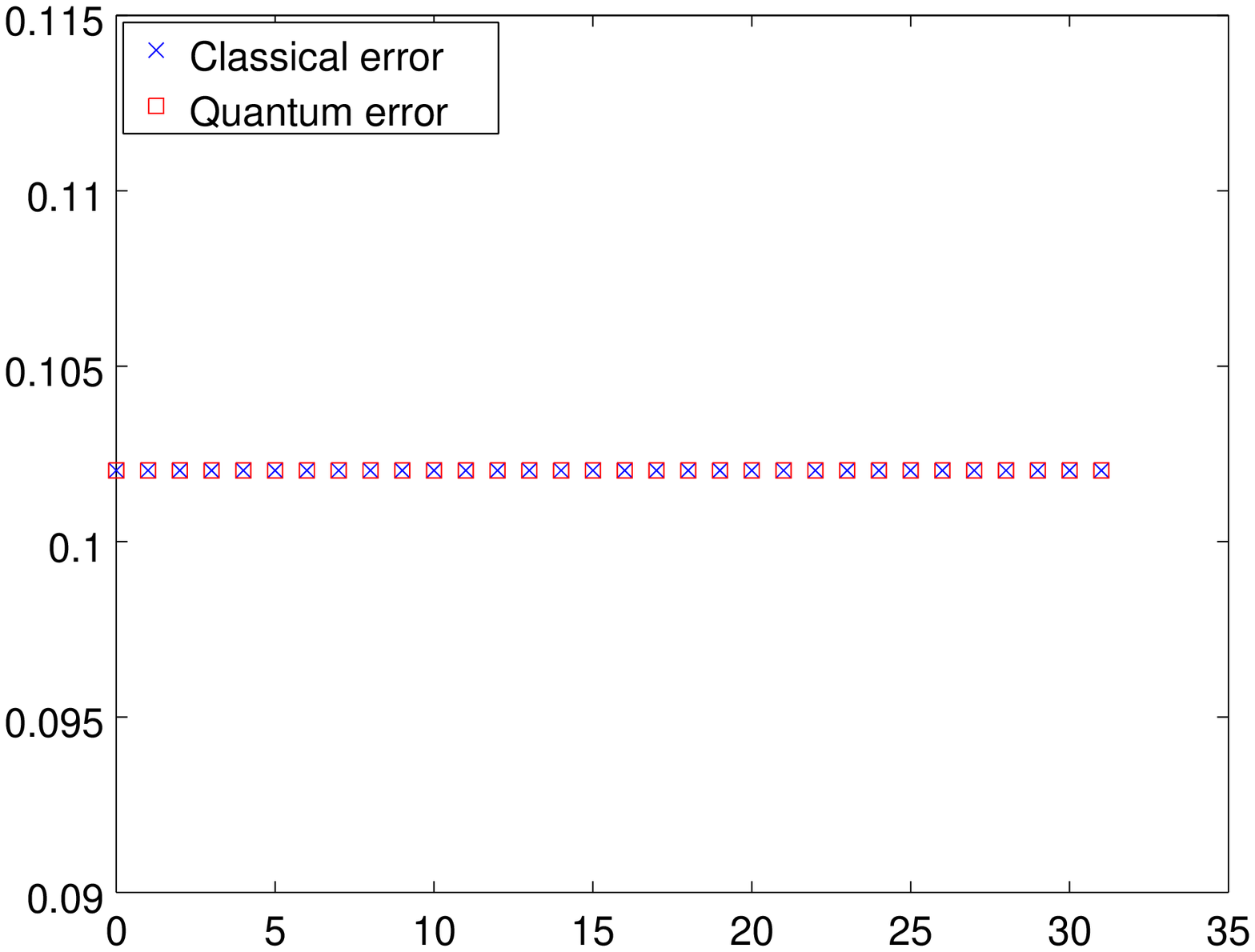}
\includegraphics[width=0.55\textwidth]{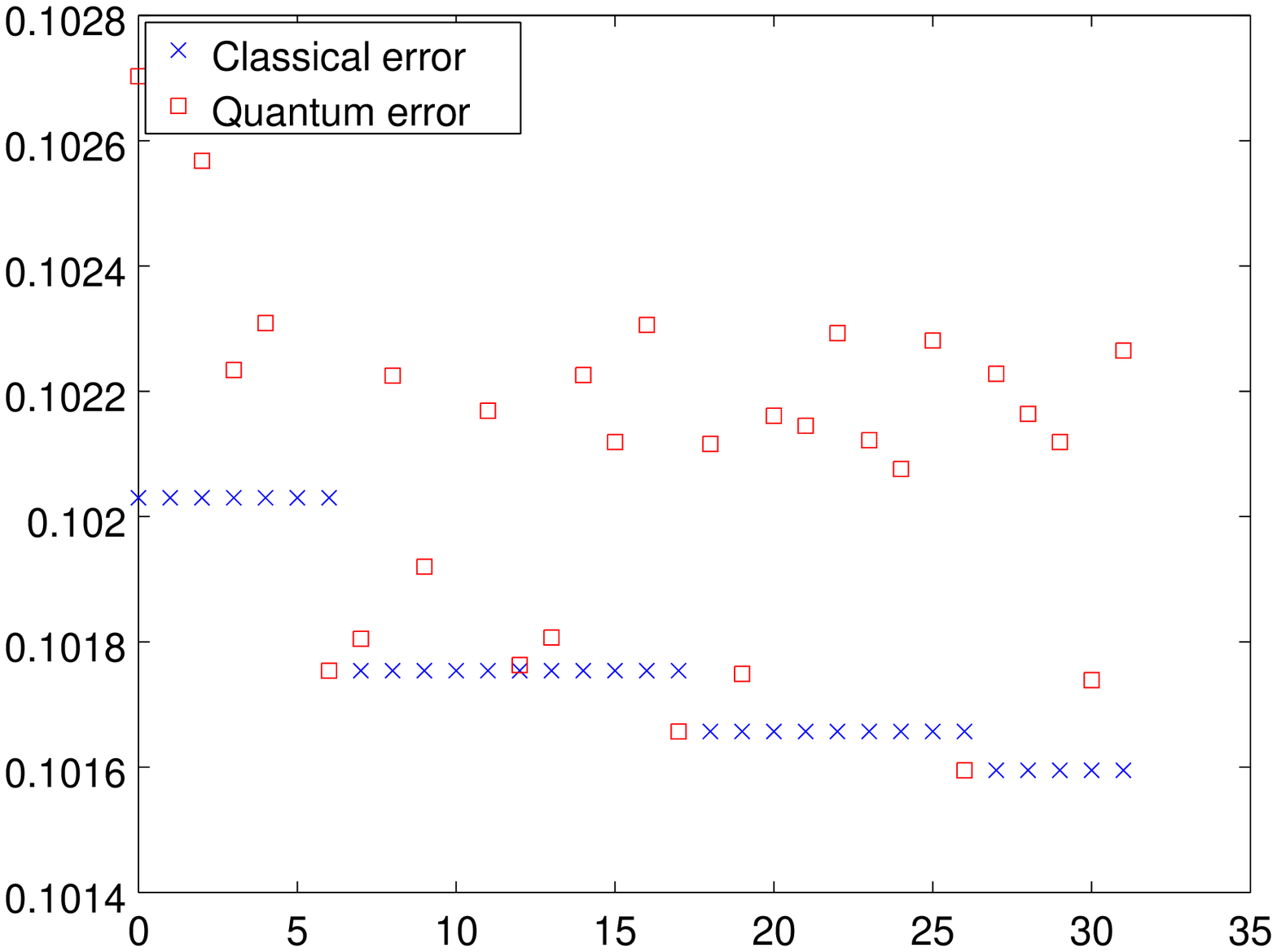}
\\
\includegraphics[width=0.55\textwidth]{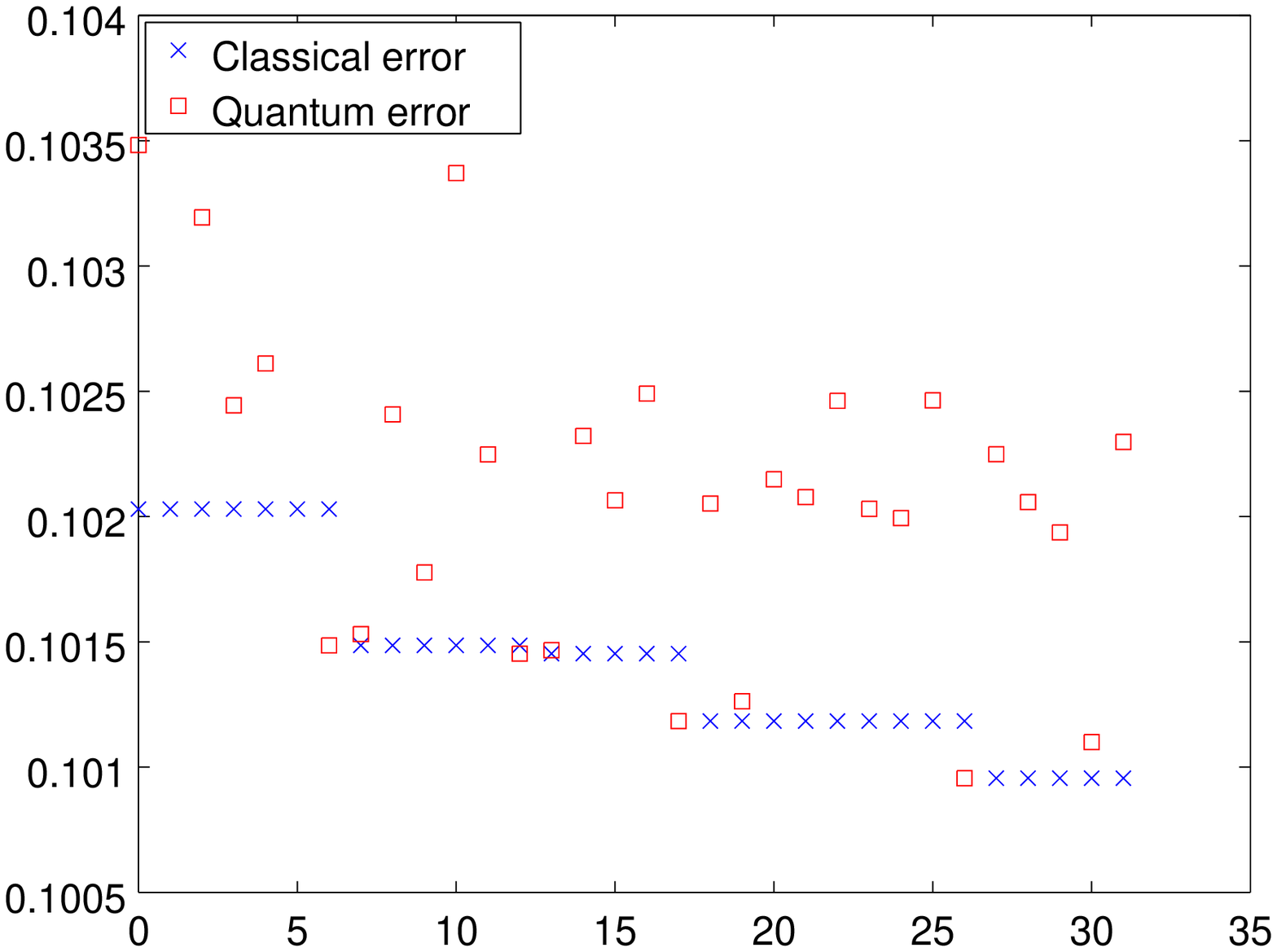}
\includegraphics[width=0.55\textwidth]{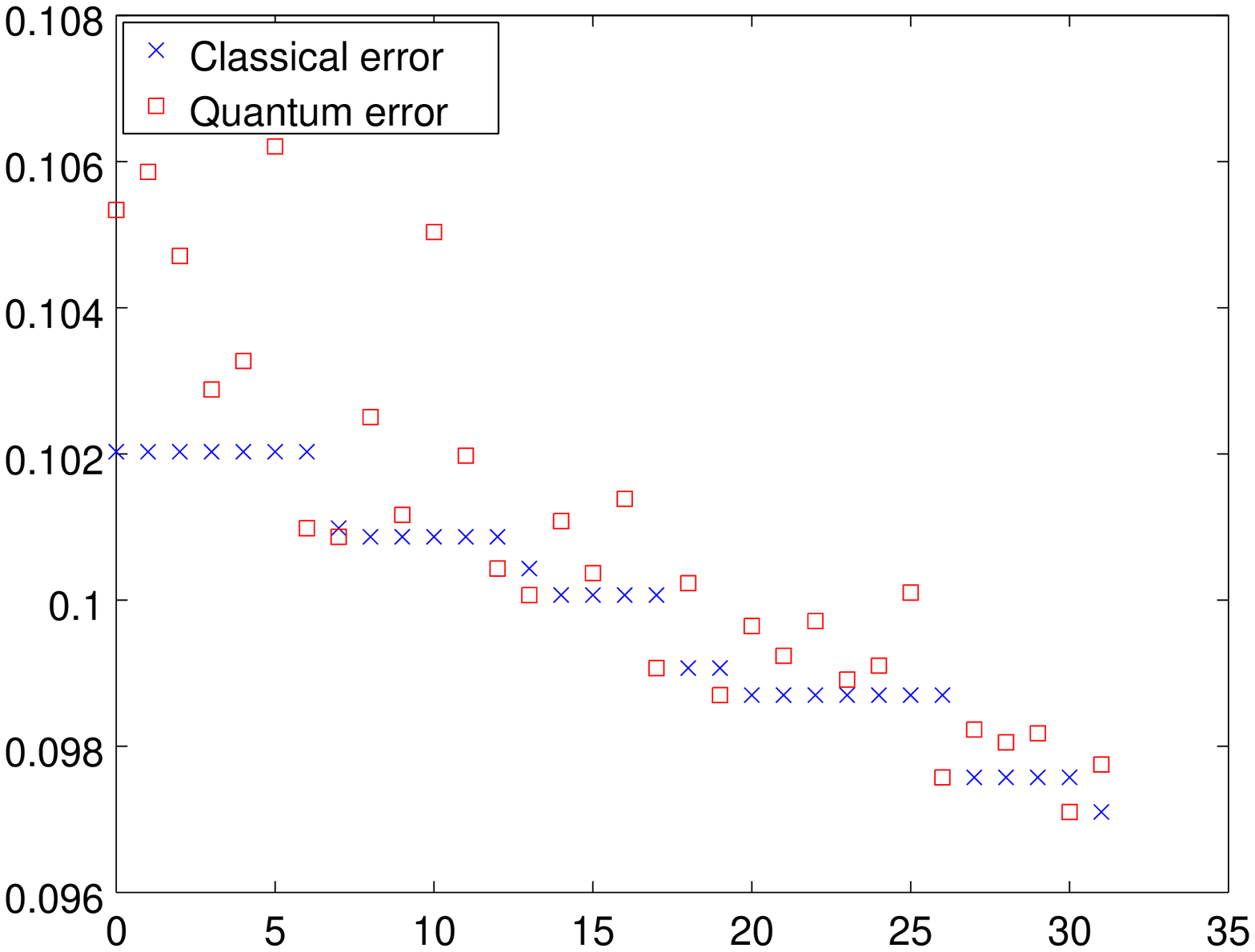}
\\
\includegraphics[width=0.55\textwidth]{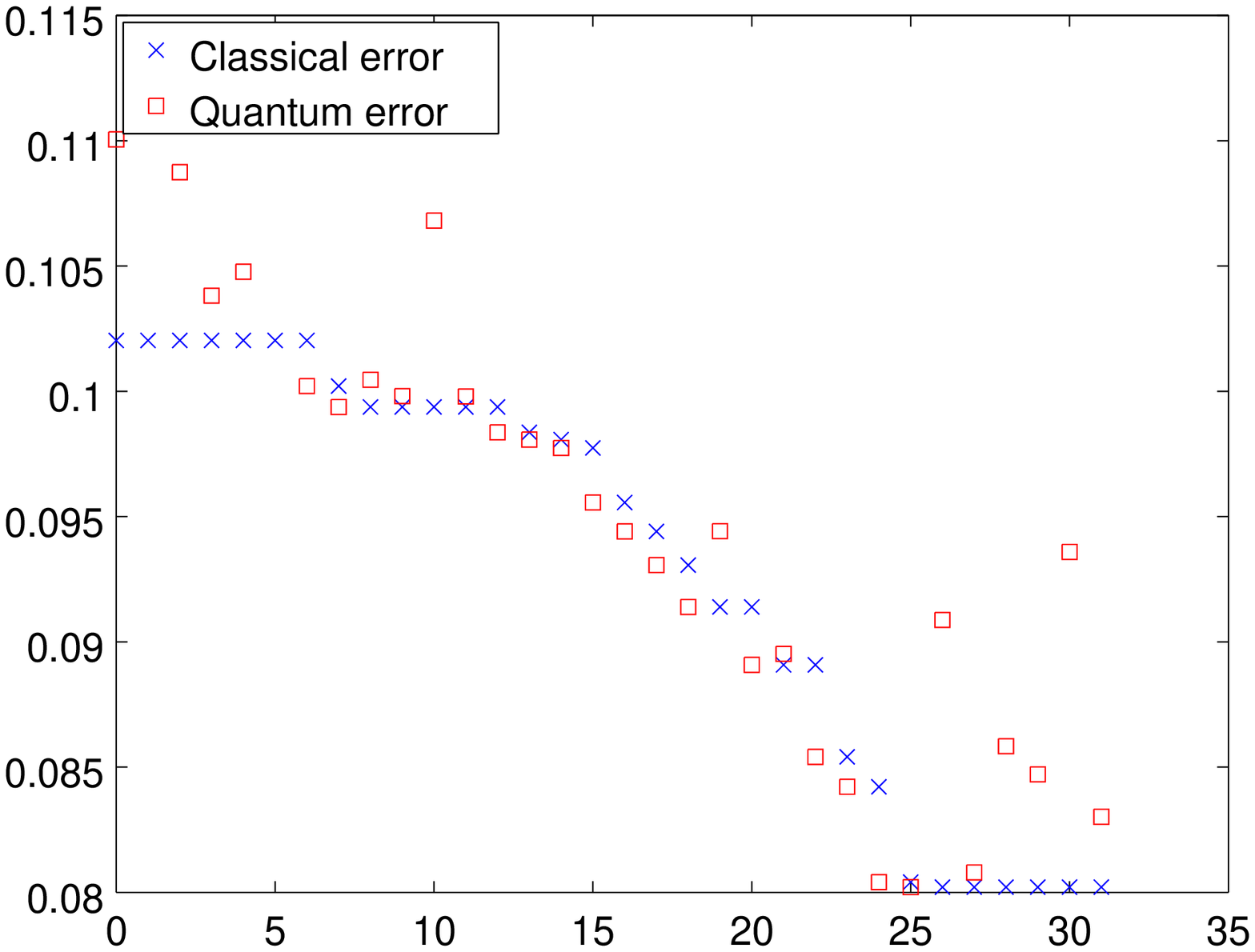}
\end{tabular}
\end{minipage}
\caption{Error decline from the neural network in the simulation of square root of a polynomial. The error from the     
classical part of the network is depicted as a (blue) $x$, while the error from the quantum side of the network is
shown as a (red) square. The upper left plot illustrates the network running with no noise for validation purposes.
The network's error with $0.5\%$ applied is depicted on the right upper plot. The middle left and right plots illustrate
the network's error in simulations with $1\%$ and $2\%$ noise. The left lowest plot indicates network's error from a test
with $4\%$ noise.}
\label{fig:errorsqrtoutput}
\end{figure}




\end{document}